\documentclass[10pt,twocolumn,letterpaper]{article}

\usepackage[pagenumbers]{cvpr} 

\definecolor{cvprblue}{rgb}{0.21,0.49,0.74}
\usepackage[pagebackref,breaklinks,colorlinks,allcolors=cvprblue]{hyperref}

\def\confName{CVPR}
\def\confYear{2027}

\usepackage{fontawesome5}

\usepackage{xspace}
\newcommand \name{DINO-WAM\xspace}
\usepackage[table]{xcolor}
\usepackage{siunitx,booktabs}
\usepackage{multirow}
\usepackage{caption}
\usepackage{enumitem}
\usepackage{hhline}
\usepackage{comment}

\usepackage{pifont}

\newcommand{\cmark}{\ding{51}}
\newcommand{\xmark}{\ding{55}}
\definecolor{dinowamblue}{RGB}{232,242,252}

\newcommand{\xxnote}[3]{}
\renewcommand{\xxnote}[3]{\color{#2}{#1: #3}}

\definecolor{dinowamred}{HTML}{C8553D}      
\definecolor{vlablue}{HTML}{7F9CC0}         
\definecolor{wamgreen}{HTML}{5FA898}        

\begin{document}

\title{DINO-WAM: What Matters for Compact World-Action Learning in Visual Feature Space}

\author{
    Davood Soleymanzadeh\textsuperscript{1} \quad Zuoxu Wang\textsuperscript{1} \quad Zhiyuan Zhang\textsuperscript{2} \quad Zhiwen Fan\textsuperscript{1} \\
    Yu She\textsuperscript{2} \quad Xiao Liang\textsuperscript{1} \quad Minghui Zheng\textsuperscript{1}\footnotemark[1] \\ [6pt]
    \normalsize $^1$Texas A\&M University \quad $^2$Purdue University \\
    {\tt mhzheng@tamu.edu} \\ [4pt]
    { \href{https://davoodsz.github.io/DINO-WAM.github.io/}{\faGlobe~Webpage}
}
}

\makeatletter
\let\@oldmaketitle\@maketitle%
\renewcommand{\@maketitle}{\@oldmaketitle%
\setcounter{figure}{0}
\centering
\begin{center}
\includegraphics[width=0.93\linewidth]{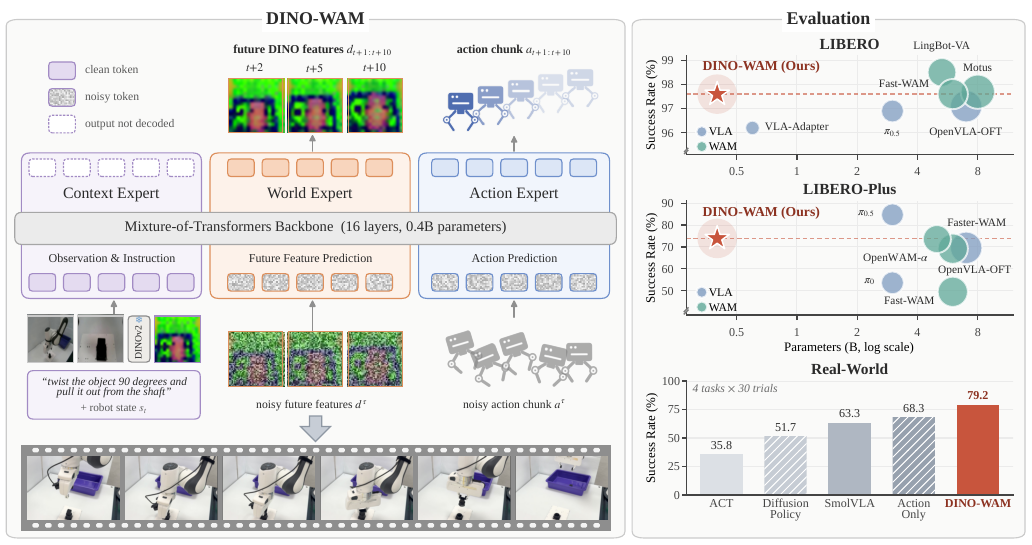}
  \captionof{figure}{\small \textbf{Method Overview.} \textit{Left}: \name simultaneously generates an action chunk and future DINOv2~\cite{oquab2023dinov2} features using a Mixture-of-Transformers~\cite{liang2024mixture}, with separate context, world, and action experts coupled through shared attention. At inference, future-feature tokens can be omitted, allowing \name to operate with the computational cost of a standard action-only policy. \textit{Right}: Despite using only 0.4B parameters and no embodied pretraining, \name (\textcolor{dinowamred}{red}) is competitive with substantially larger VLAs (\textcolor{vlablue}{blue}), including OpenVLA-OFT~\cite{kim2025fine} and $\pi_{0.5}$~\cite{intelligence2025pi_}, as well as world-action models (\textcolor{wamgreen}{green}), including Motus~\cite{bi2026motus}, LingBot-VA~\cite{li2026causal}, and Fast-WAM~\cite{yuan2026fast}, on both LIBERO~\cite{liu2023libero} and zero-shot LIBERO-Plus~\cite{fei2025libero}, while also achieving the highest average success rate on the considered real-world disassembly tasks compared with the evaluated policy baselines.}
  \label{fig:openning}
  \end{center}
  \vspace{0.1in}
  }
\makeatother
\maketitle

\renewcommand{\thefootnote}{\fnsymbol{footnote}}
\footnotetext[1]{Corresponding Author.}
\renewcommand{\thefootnote}{\arabic{footnote}}
\begin{abstract}
World-action models (WAMs) improve robot policies by augmenting action learning with future prediction, yet it remains unclear which design choices matter when future dynamics are modeled in pretrained visual feature spaces. We study this question through \textbf{\name}, a compact WAM that simultaneously learns robot actions and future visual features using flow matching and a structured attention mask that keeps future prediction auxiliary to action generation. We systematically evaluate visual representation, context features, future-target viewpoint, target parameterization, temporal density, and action-world coupling. Our results show that masked future-feature supervision is more effective than direct joint generation, while absolute targets and dense near-term supervision provide the strongest performance. Despite using only 0.4B parameters and no embodied pretraining, \textbf{\name} achieves 97.6\% average success on LIBERO, 74.4\% on zero-shot LIBERO-Plus, and improves real-world disassembly success from 68.3\% with an action-only baseline to 79.2\%.
\end{abstract}    
\section{Introduction} \label{sec:intro}
Vision-Language-Action (VLA) models have emerged as a promising framework for general-purpose robotic manipulation, mapping language instructions and visual observations to robot actions across diverse objects, tasks, and environments~\cite{black2024pi_0, intelligence2025pi_, kim2024openvla, kim2025fine}. Most existing VLAs, however, are trained primarily through action prediction, directly mapping the current task context to an action sequence~\cite{black2024pi_0, kim2024openvla}. Consequently,  the policy must learn not only how to act, but also the visual and temporal structure required for control from action supervision alone. This can be particularly challenging for long-horizon and contact-rich tasks, where successful control depends on understanding how the scene evolves as actions are executed. Without explicit supervision on such temporal evolution, these reactive policies may capture task-relevant correlations while learning only limited representations of the underlying environment dynamics~\cite{li2026causal, ye2026world}.

Robot control often requires anticipating how the scene will evolve under action. World-action models (WAMs) address this by coupling action generation with prediction of future observations or latent visual states~\cite{li2026causal, ye2026world, li2026efficient, li2026light, huang2026long}. This predictive objective provides an explicit temporal learning signal for modeling interaction dynamics. However, many existing WAMs build on large video-generation models, introducing substantial computational overhead during training and inference~\cite{li2026light, li2026efficient}. Moreover, their latent spaces are often designed around visual reconstruction or generation, potentially allocating capacity to appearance details that are less important for control~\cite{jiang2026rethinking}.

Recent studies suggest that photorealistic future generation is not necessary for effective world-action learning. Fast-WAM~\cite{yuan2026fast} shows that explicit future generation can be removed at inference with limited performance degradation, while Efficient-WAM~\cite{li2026efficient} reduces future prediction cost through video-side compression. More directly, ReWAM~\cite{jiang2026rethinking} finds that pretrained DINO features can provide stronger prediction targets than Video-VAE latents despite lower reconstruction fidelity, and LDA~\cite{lyu2602lda} similarly models future robot dynamics in a structured DINO feature space. Concurrent OpenWAM studies investigate complementary WAM design choices: one explores composable joint, decoupled, and directional world-action generation~\cite{yu2026openwam}, while another systematically studies generative priors, information flow, synchronized denoising, and embodied pretraining~\cite{wang2026openwam}. Together, these works motivate a shift from asking whether a policy should predict the future to asking how future prediction should be incorporated into policy learning.

Motivated by these observations, we introduce \name, a compact world-action model that simultaneously predicts robot actions and future frozen DINOv2 features~\cite{oquab2023dinov2}, as illustrated in Figure~\ref{fig:openning}. Its masked formulation separates action generation from the world stream, allowing future prediction to act as an auxiliary supervision without requiring future-feature generation at deployment. We examine visual representation, context features, future-target viewpoint, feature parameterization, temporal placement and density, and action-world information flow. Our evaluations show that these choices affect policy performance: masked future-feature supervision outperforms direct joint generation, while DINOv2 features, absolute targets, dense near-term supervision, and multi-view future targets provide the strongest results for the masked policy.

\begin{itemize}[leftmargin=*,itemsep=2pt,topsep=4pt]

    \item We introduce \name, a compact feature-space world-action model that uses frozen DINOv2 representations and a masked action-world architecture for robotic manipulation. We characterize the effects of visual representation, context construction, future-target viewpoint, target parameterization, temporal supervision, and action-world coupling.

    \item Through controlled simulation and real-world experiments, we show that the effectiveness of future prediction depends on how it is represented and coupled to action learning, with masked future-feature supervision providing consistent gains over action-only and directly coupled alternatives.
\end{itemize}

\section{Related Work} \label{sec:relatedworks}
\noindent
\textbf{Vision-Language-Action Models.} Vision-Language-Action (VLA) models map visual observations and language instructions to robot actions using large pretrained vision-language backbones~\cite{black2024pi_0, intelligence2025pi_, kim2024openvla, kim2025fine, bjorck2025gr00t, bu2025agibot, team2025gemini, jiang2026galaxy, wen2025dexvla, soleymanzadeh2026phr}. By transferring semantic knowledge from large-scale pretraining, these models achieve strong generalization across tasks, objects, and environments. Most VLAs, however, are optimized primarily for action prediction and provide limited explicit supervision on how visual states evolve during interaction~\cite{li2026causal, ye2026world}. \name is complementary to this line of work: it retains direct action generation at deployment while augmenting policy learning with prediction of future visual features in a fixed pretrained representation space.

\begin{figure*}[t]
  \centering
  \includegraphics[width=0.95\linewidth]{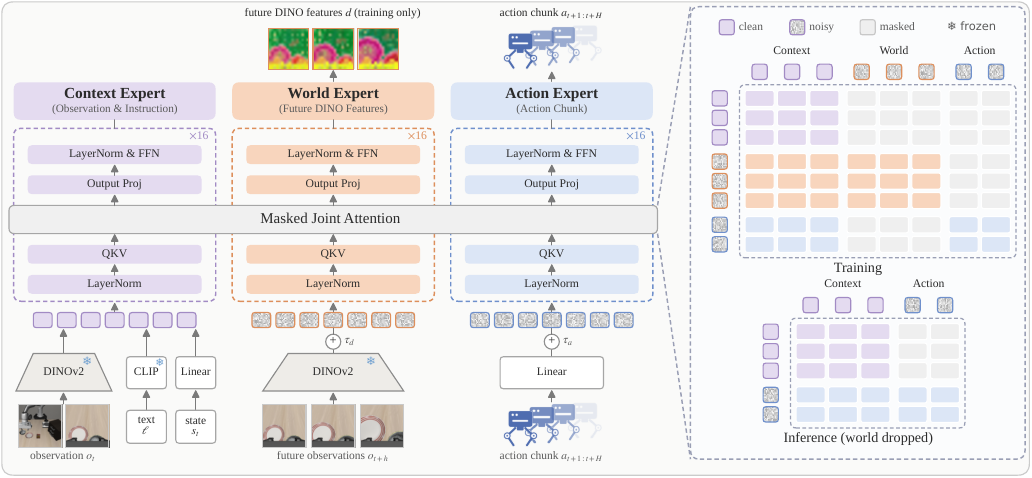}
  \caption{\small \textbf{\name Architecture and Attention Mask.} \textit{Left}: \name adopts a Mixture-of-Transformers~\cite{liang2024mixture} with three modality-specific experts, each with its own normalization, projection, and feed-forward layers, while sharing masked joint attention. The \emph{context expert} encodes the current camera views using a frozen DINOv2~\cite{oquab2023dinov2} encoder, the language instruction using a frozen CLIP text encoder~\cite{radford2021learning}, and the robot state through a linear projection. The \emph{world expert} denoises future DINOv2 features $d$, while the \emph{action expert} denoises the action chunk $a_{t+1:t+H}$. The two streams are independently corrupted under flow matching at flow times $\tau_d$ and $\tau_a$. \textit{Right}: The attention mask used during training. World and action tokens attend to the shared context and within their own streams, but are prevented from attending to one another. This allows future-feature prediction to shape the shared context representation without allowing action generation to depend on predicted future features. At inference, the world tokens are removed and only the action chunk is generated, retaining the computational cost of a standard action-only policy.}
  \label{fig:framework}
\end{figure*}

\vspace{0.2cm}
\noindent
\textbf{World Action Models.} World Action Models (WAMs) augment robot policies with prediction of future observations, providing an explicit learning signal for environment dynamics~\cite{ye2026world, li2026causal}. Existing methods differ in how future prediction is coupled with action generation and the representation used to model the future. Some approaches jointly generate future visual states and robot actions~\cite{ye2026world, yan2026flex, zhu2025unified}, while others use future prediction as an auxiliary supervision and retain direct action prediction at inference time~\cite{yuan2026fast}. Concurrent OpenWAM studies explore complementary design choices: one investigates composable joint, decoupled, and directional world-action generation~\cite{yu2026openwam}, while another studies generative priors, information flow, synchronized denoising, and embodied pretraining~\cite{wang2026openwam}.

Early WAMs operate in RGB space~\cite{du2023learning, hu2024video} or video-VAE latents~\cite{yuan2026fast, ye2026world, kim2026cosmos, zhu2025unified, pai2025mimic, li2026causal}, often building on pretrained video-generation models. Recent work has shifted toward semantic feature spaces less tied to pixel reconstruction. ReWAM~\cite{jiang2026rethinking} shows that pretrained DINO features can provide effective future-prediction targets despite limited reconstruction fidelity, while LDA~\cite{lyu2602lda} models future robot dynamics directly in a structured DINO representation. Other approaches similarly investigate semantic or learned latent spaces for predictive control~\cite{deng2026dexworldmodel, wang2026repwam, yang2026lila}, while broader studies examine how reconstruction-oriented and semantic representations influence prediction and control~\cite{jha2026reconstruction, yeom2026makes}.

Our work builds on both directions but addresses a complementary question: \emph{what makes future supervision useful when the future is represented in a frozen visual feature space?} Specifically, we examine visual encoder choice, context representation, future-target viewpoint, target parameterization, temporal density, and action-world coupling. This controlled design allows us to distinguish the effect of the future representation itself from how that supervision is incorporated into action learning.

\setlength{\tabcolsep}{3pt}
\begin{table*}[htbp]
\begin{center}
\captionof{table}{\small \textbf{Results on LIBERO.} Success rates are reported across four task suites for models with (\cmark) and without (\xmark) embodied pretraining (Embodied PT.). Best results are \textbf{bold}, and second-best results are \underline{underlined}.}
\label{tab:libero}
\resizebox{0.85\textwidth}{!}{%
\begin{tabular}
{>{\raggedright\arraybackslash}p{0.1\textwidth}>{\raggedright\arraybackslash}p{0.17\textwidth}>{\raggedright\arraybackslash}p{0.07\textwidth}>{\raggedright\arraybackslash}p{0.14\textwidth}>{\raggedright\arraybackslash}p{0.1\textwidth}>{\raggedright\arraybackslash}p{0.08\textwidth}>{\raggedright\arraybackslash}p{0.08\textwidth}>{\raggedright\arraybackslash}p{0.08\textwidth}>{\raggedright\arraybackslash}p{0.1\textwidth}}
\toprule
\rowcolor{gray!20} \multicolumn{1}{l}{\textbf{Type}}&\multicolumn{1}{l}{\textbf{Method}}&\multicolumn{1}{|c}{\textbf{Params}}&\multicolumn{1}{c}{\textbf{Embodied PT.}}&\multicolumn{1}{|c}{\textbf{Spatial} $\uparrow$}&\multicolumn{1}{c}{\textbf{Object} $\uparrow$}&\multicolumn{1}{c}{\textbf{Goal} $\uparrow$}&\multicolumn{1}{c}{\textbf{Long} $\uparrow$}& \multicolumn{1}{|c}{\textbf{Average} $\uparrow$} \\
\midrule
\multirow{7}{*}{\small VLA} & \small OpenVLA~\cite{kim2024openvla}&\multicolumn{1}{|c}{\small 7B}&\multicolumn{1}{c|}{\small \cmark}&\multicolumn{1}{c}{\small 84.7\%}&\multicolumn{1}{c}{\small 88.4\%}&\multicolumn{1}{c}{\small 79.2\%}&\multicolumn{1}{c}{\small 53.7\%}&\multicolumn{1}{|c}{\small 76.5\%} \\
&\small OpenVLA-OFT~\cite{kim2025fine}&\multicolumn{1}{|c}{\small 7B}&\multicolumn{1}{c|}{\small \cmark}&\multicolumn{1}{c}{\small 97.6\%}&\multicolumn{1}{c}{\small 98.4\%}&\multicolumn{1}{c}{\small \underline{97.9}\%}&\multicolumn{1}{c}{\small 94.5\%}&\multicolumn{1}{|c}{\small 97.1\%} \\
&\small $\pi_0$~\cite{black2024pi_0}&\multicolumn{1}{|c}{\small 3B}&\multicolumn{1}{c|}{\small \cmark}&\multicolumn{1}{c}{\small 96.8\%}&\multicolumn{1}{c}{\small 98.8\%}&\multicolumn{1}{c}{\small 95.8\%}&\multicolumn{1}{c}{\small 85.2\%}&\multicolumn{1}{|c}{\small 94.1\%} \\
&\small $\pi_{0.5}$~\cite{intelligence2025pi_}&\multicolumn{1}{|c}{\small 3B}&\multicolumn{1}{c|}{\small \cmark}&\multicolumn{1}{c}{\small \textbf{98.8}\%}&\multicolumn{1}{c}{\small 98.2\%}&\multicolumn{1}{c}{\small \textbf{98.0}\%}&\multicolumn{1}{c}{\small 92.4\%}&\multicolumn{1}{|c}{\small 96.9\%}\\
&\small VLA-Adapter~\cite{wang2026vla}&\multicolumn{1}{|c}{\small 0.6B}&\multicolumn{1}{c|}{\small \xmark}&\multicolumn{1}{c}{\small 96.0\%}&\multicolumn{1}{c}{\small 96.8\%}&\multicolumn{1}{c}{\small 97.4\%}&\multicolumn{1}{c}{\small 94.4\%}&\multicolumn{1}{|c}{\small 96.2\%} \\
&\small StarVLA~\cite{community2026starvla}&\multicolumn{1}{|c}{4B}&\multicolumn{1}{c|}{\small \xmark}&\multicolumn{1}{c}{\small 97.8\%}&\multicolumn{1}{c}{\small 98.6\%}&\multicolumn{1}{c}{\small 96.2\%}&\multicolumn{1}{c}{\small 93.8\%}&\multicolumn{1}{|c}{\small 96.6\%} \\
&\small GR00T-N1.6~\cite{bjorck2025gr00t}&\multicolumn{1}{|c}{3.3B}&\multicolumn{1}{c|}{\small \cmark}&\multicolumn{1}{c}{\small 97.7\%}&\multicolumn{1}{c}{\small 98.5\%}&\multicolumn{1}{c}{\small 97.5\%}&\multicolumn{1}{c}{\small 94.4\%}&\multicolumn{1}{|c}{\small 97.0\%} \\
\midrule
\multirow{5}{*}{\small WAM}&\small Motus~\cite{bi2026motus}&\multicolumn{1}{|c}{\small 8B}&\multicolumn{1}{c|}{\small \cmark}&\multicolumn{1}{c}{\small 96.8\%}&\multicolumn{1}{c}{\small \underline{99.8}\%}&\multicolumn{1}{c}{\small 96.6\%}&\multicolumn{1}{c}{\small \underline{97.6}\%}&\multicolumn{1}{|c}{\small \underline{97.7}\%} \\
&\small LingBot-VA~\cite{li2026causal}&\multicolumn{1}{|c}{\small 5.3B}&\multicolumn{1}{c|}{\small \cmark}&\multicolumn{1}{c}{\small \underline{98.5}\%}&\multicolumn{1}{c}{\small 99.6\%}&\multicolumn{1}{c}{\small 97.2\%}&\multicolumn{1}{c}{\small \textbf{98.5}\%}&\multicolumn{1}{|c}{\small \textbf{98.5}\%} \\
&\small Fast-WAM~\cite{yuan2026fast}&\multicolumn{1}{|c}{\small 6B}&\multicolumn{1}{c|}{\small \xmark}&\multicolumn{1}{c}{\small 98.2\%}&\multicolumn{1}{c}{\small \textbf{100.0}\%}&\multicolumn{1}{c}{\small 97.0\%}&\multicolumn{1}{c}{\small 95.2\%}&\multicolumn{1}{|c}{\small 97.6\%} \\
&\small Light-WAM~\cite{li2026light}&\multicolumn{1}{|c}{\small 2B}&\multicolumn{1}{c|}{\small \xmark}&\multicolumn{1}{c}{\small 98.2\%}&\multicolumn{1}{c}{\small 99.6\%}&\multicolumn{1}{c}{\small 97.8\%}&\multicolumn{1}{c}{\small 93.0\%}&\multicolumn{1}{|c}{\small 97.2\%} \\
\rowcolor{dinowamblue} & \fontsize{8}{9}\selectfont\bfseries \name&\multicolumn{1}{|c}{\small 0.4B}&\multicolumn{1}{c|}{\small \xmark}&\multicolumn{1}{c}{\small 98.0\%}&\multicolumn{1}{c}{\small 99.6\%}&\multicolumn{1}{c}{\small 97.4\%}&\multicolumn{1}{c}{\small 95.4\%}&\multicolumn{1}{|c}{\small 97.6\%} \\
\bottomrule
\end{tabular}}
\end{center}
\end{table*}

\setlength{\tabcolsep}{3pt}
\begin{table*}[htbp]
\begin{center}
\captionof{table}{\small \textbf{Zero-shot Results on LIBERO-Plus.} Success rates are reported under seven types of perturbations for models with (\cmark) and without (\xmark) embodied pretraining (Embodied PT.). ``Lang.'' denotes Language, and ``Backg.'' denotes Background. Best results are \textbf{bold}, and second-best results are \underline{underlined}.}
\label{tab:liberoplus}
\resizebox{1.0\textwidth}{!}{%
\begin{tabular}
{>{\raggedright\arraybackslash}p{0.1\textwidth}>{\raggedright\arraybackslash}p{0.17\textwidth}>{\raggedright\arraybackslash}p{0.07\textwidth}>{\raggedright\arraybackslash}p{0.14\textwidth}>{\raggedright\arraybackslash}p{0.1\textwidth}>{\raggedright\arraybackslash}p{0.08\textwidth}>{\raggedright\arraybackslash}p{0.08\textwidth}>{\raggedright\arraybackslash}p{0.08\textwidth}>{\raggedright\arraybackslash}p{0.08\textwidth}>{\raggedright\arraybackslash}p{0.08\textwidth}>{\raggedright\arraybackslash}p{0.08\textwidth}>{\raggedright\arraybackslash}p{0.1\textwidth}}
\toprule
\rowcolor{gray!20} \multicolumn{1}{l}{\textbf{Type}}&\multicolumn{1}{l}{\textbf{Method}}&\multicolumn{1}{|c}{\textbf{Params}}&\multicolumn{1}{c}{\textbf{Embodied PT.}}&\multicolumn{1}{|c}{\textbf{Camera} $\uparrow$}&\multicolumn{1}{c}{\textbf{Robot} $\uparrow$}&\multicolumn{1}{c}{\textbf{Lang.} $\uparrow$}&\multicolumn{1}{c}{\textbf{Light} $\uparrow$}&\multicolumn{1}{c}{\textbf{Backg.} $\uparrow$}&\multicolumn{1}{c}{\textbf{Noise} $\uparrow$}&\multicolumn{1}{c}{\textbf{Layout} $\uparrow$}& \multicolumn{1}{|c}{\textbf{Average} $\uparrow$} \\
\midrule
\multirow{7}{*}{\small VLA}&\small OpenVLA~\cite{kim2024openvla}&\multicolumn{1}{|c}{\small 7B}&\multicolumn{1}{c|}{\small \cmark}&\multicolumn{1}{c}{\small 0.6\%}&\multicolumn{1}{c}{\small 3.9\%}&\multicolumn{1}{c}{\small 38.7\%}&\multicolumn{1}{c}{\small 14.2\%}&\multicolumn{1}{c}{\small 38.4\%}&\multicolumn{1}{c}{\small 5.4\%}&\multicolumn{1}{c}{\small 42.5\%}&\multicolumn{1}{|c}{\small 19.7\%} \\
&\small OpenVLA-OFT~\cite{kim2025fine}&\multicolumn{1}{|c}{\small 7B}&\multicolumn{1}{c|}{\small \cmark}&\multicolumn{1}{c}{\small 56.4\%}&\multicolumn{1}{c}{\small 31.9\%}&\multicolumn{1}{c}{\small 79.4\%}&\multicolumn{1}{c}{\small 88.7\%}&\multicolumn{1}{c}{\small \underline{93.3}\%}&\multicolumn{1}{c}{\small 75.8\%}&\multicolumn{1}{c}{\small 74.4\%}&\multicolumn{1}{|c}{\small 69.6\%} \\
&\small $\pi_0$~\cite{black2024pi_0}&\multicolumn{1}{|c}{\small 3B}&\multicolumn{1}{c|}{\small \cmark}&\multicolumn{1}{c}{\small 13.8\%}&\multicolumn{1}{c}{\small 6.0\%}&\multicolumn{1}{c}{\small 58.8\%}&\multicolumn{1}{c}{\small 85.0\%}&\multicolumn{1}{c}{\small 81.4\%}&\multicolumn{1}{c}{\small 79.0\%}&\multicolumn{1}{c}{\small 68.9\%}&\multicolumn{1}{|c}{\small 53.6\%} \\
&\small $\pi_{0.5}$~\cite{intelligence2025pi_}&\multicolumn{1}{|c}{\small 3B}&\multicolumn{1}{c|}{\small \cmark}&\multicolumn{1}{c}{\small \underline{72.2}\%}&\multicolumn{1}{c}{\small \textbf{76.8}\%}&\multicolumn{1}{c}{\small 85.1\%}&\multicolumn{1}{c}{\small 96.2\%}&\multicolumn{1}{c}{\small \textbf{95.5}\%}&\multicolumn{1}{c}{\small \textbf{87.0}\%}&\multicolumn{1}{c}{\small \textbf{89.6}\%}&\multicolumn{1}{|c}{\small \textbf{84.7}\%} \\
&\small $\pi_{0}$-Fast~\cite{pertsch2025fast}&\multicolumn{1}{|c}{\small 3.5B}&\multicolumn{1}{c|}{\small \cmark}&\multicolumn{1}{c}{\small 65.1\%}&\multicolumn{1}{c}{\small 21.6\%}&\multicolumn{1}{c}{\small 61.0\%}&\multicolumn{1}{c}{\small 73.2\%}&\multicolumn{1}{c}{\small 73.2\%}&\multicolumn{1}{c}{\small 74.4\%}&\multicolumn{1}{c}{\small 68.8\%}&\multicolumn{1}{|c}{\small 61.6\%} \\
&\small WorldVLA~\cite{cen2025worldvla}&\multicolumn{1}{|c}{\small 7B}&\multicolumn{1}{c|}{\small \cmark}&\multicolumn{1}{c}{\small 0.1\%}&\multicolumn{1}{c}{\small 27.9\%}&\multicolumn{1}{c}{\small 41.6\%}&\multicolumn{1}{c}{\small 43.7\%}&\multicolumn{1}{c}{\small 17.1\%}&\multicolumn{1}{c}{\small 10.9\%}&\multicolumn{1}{c}{\small 38.0\%}&\multicolumn{1}{|c}{\small 25.0\%} \\
&\small UniVLA~\cite{bu2025univla}&\multicolumn{1}{|c}{\small 7B}&\multicolumn{1}{c|}{\small \cmark}&\multicolumn{1}{c}{\small 1.8\%}&\multicolumn{1}{c}{\small 46.2\%}&\multicolumn{1}{c}{\small 69.6\%}&\multicolumn{1}{c}{\small 69.0\%}&\multicolumn{1}{c}{\small 81.0\%}&\multicolumn{1}{c}{\small 21.2\%}&\multicolumn{1}{c}{\small 31.9\%}&\multicolumn{1}{|c}{\small 42.9\%} \\
\midrule
\multirow{5}{*}{\small WAM}&\small Fast-WAM~\cite{yuan2026fast}&\multicolumn{1}{|c}{\small 6B}&\multicolumn{1}{c|}{\small \xmark}&\multicolumn{1}{c}{\small 15.4\%}&\multicolumn{1}{c}{\small 44.4\%}&\multicolumn{1}{c}{\small 68.5\%}&\multicolumn{1}{c}{\small 79.0\%}&\multicolumn{1}{c}{\small 52.6\%}&\multicolumn{1}{c}{\small 37.3\%}&\multicolumn{1}{c}{\small 61.5\%}&\multicolumn{1}{|c}{\small 49.5\%} \\
&\small OpenWAM-$\alpha$~\cite{wang2026openwam}&\multicolumn{1}{|c}{\small 6B}&\multicolumn{1}{c|}{\small \xmark}&\multicolumn{1}{c}{\small 33.8\%}&\multicolumn{1}{c}{\small \underline{76.1}\%}&\multicolumn{1}{c}{\small \underline{88.0}\%}&\multicolumn{1}{c}{\small \textbf{97.0}\%}&\multicolumn{1}{c}{\small 87.1\%}&\multicolumn{1}{c}{\small 39.8\%}&\multicolumn{1}{c}{\small 77.5\%}&\multicolumn{1}{|c}{\small 69.2\%} \\
&\small Faster-WAM~\cite{zhao2026faster}&\multicolumn{1}{|c}{\small 5B}&\multicolumn{1}{c|}{\small \xmark}&\multicolumn{1}{c}{\small 53.8\%}&\multicolumn{1}{c}{\small 71.6\%}&\multicolumn{1}{c}{\small \textbf{94.7}\%}&\multicolumn{1}{c}{\small \underline{96.3}\%}&\multicolumn{1}{c}{\small 61.3\%}&\multicolumn{1}{c}{\small 63.6\%}&\multicolumn{1}{c}{\small \underline{79.1}\%}&\multicolumn{1}{|c}{\small 73.6\%} \\
&\small BridgeWAM~\cite{fang2026bridgewam}&\multicolumn{1}{|c}{\small 5B}&\multicolumn{1}{c|}{\small \xmark}&\multicolumn{1}{c}{\small 56.5\%}&\multicolumn{1}{c}{\small 60.5\%}&\multicolumn{1}{c}{\small 85.1\%}&\multicolumn{1}{c}{\small 92.2\%}&\multicolumn{1}{c}{\small 69.2\%}&\multicolumn{1}{c}{\small 73.6\%}&\multicolumn{1}{c}{\small 75.7\%}&\multicolumn{1}{|c}{\small 72.2\%} \\
\midrule
\rowcolor{dinowamblue} &\small \textbf{\name}&\multicolumn{1}{|c}{\small 0.4B}&\multicolumn{1}{c|}{\small \xmark}&\multicolumn{1}{c}{\small \textbf{87.3}\%}&\multicolumn{1}{c}{\small 36.3\%}&\multicolumn{1}{c}{\small 79.2\%}&\multicolumn{1}{c}{\small 93.1\%}&\multicolumn{1}{c}{\small 65.7\%}&\multicolumn{1}{c}{\small \underline{83.3}\%}&\multicolumn{1}{c}{\small 77.6\%}&\multicolumn{1}{|c}{\small \underline{74.4}\%} \\
\bottomrule
\end{tabular}}
\end{center}
\end{table*}

\section{\name} \label{sec:DINO-WAM}
We present \name, a world-action model that simultaneously models robot actions and future visual representations in the feature space of a frozen DINOv2 encoder~\cite{oquab2023dinov2}. Figure~\ref{fig:framework} illustrates the \name architecture and its attention mask. 

\vspace{0.2cm}
\noindent
\textbf{Problem Statement.} At time $t$, the robot receives visual observations $o_t$, proprioceptive state $s_t$, and language instruction $l$. A frozen DINOv2 encoder~\cite{oquab2023dinov2} extracts visual features from the current observations, which are fused and spatially pooled into a compact representation $d_t = P(f_{\mathrm{DINO}}(o_t))$. Given the current task context, an action chunk $a_{t+1:t+H}$ together with future visual representations $d_{t+1:t+H}$ are modeled:
\begin{equation}
p_\theta \left(a_{t+1:t+H}, d_{t+1:t+H} \mid d_t,s_t,l \right).
\end{equation}

\vspace{0.2cm}
\noindent
\textbf{Flow Matching.} Both action and future-feature generation are formulated using flow matching~\cite{lipman2022flow}. Given a clean sample $x^0 \sim p_0$ and Gaussian noise $\epsilon \sim \mathcal{N}(0,I)$, the conditional optimal transport path~\cite{liu2022flow} is:
\begin{equation}
    x^\tau = (1-\tau)x^0 + \tau\epsilon,
    \qquad \tau \in [0,1],
\end{equation}
which interpolates from clean data at $\tau=0$ to Gaussian noise at $\tau=1$. The corresponding conditional vector field is
\begin{equation}
    u_\tau(x^\tau \mid x^0) = \frac{d x^\tau}{d\tau} = \epsilon - x^0.
\end{equation}
The marginal vector field associated with this probability path is
\begin{equation}
    u_\tau(x^\tau)
    =
    \mathbb{E}_{p(x^0\mid x^\tau)}
    \left[
        u_\tau(x^\tau\mid x^0)
    \right].
\end{equation}
Since this marginal field is generally intractable, a neural vector field $v_\theta(x^\tau,\tau)$ is trained by regressing to the tractable conditional field:
\begin{equation}
    \mathcal{L}_{\mathrm{FM}} = \mathbb{E}_{\substack{x^0\sim p_0,\,\epsilon\sim\mathcal{N}(0,I),\\ \tau\sim\mathcal{U}(0,1)}} \left[ \left\| v_\theta(x^\tau,\tau) - (\epsilon-x^0) \right\|_2^2 \right].
\end{equation}
At inference, generation starts from $\hat{x}^{1}\sim\mathcal{N}(0,I)$ and integrates the learned vector field backward from $\tau=1$ to $\tau=0$:
\begin{equation}
    \hat{x}^{0}
    =
    \hat{x}^{1}
    +
    \int_{1}^{0}
    v_\theta(\hat{x}^{\tau},\tau)\,d\tau.
\end{equation}

\vspace{0.2cm}
\noindent
\textbf{Action and Future-Feature Modeling.} Future-feature prediction is utilized to supervise policy learning without requiring future generation at inference. Let $a^0=a_{t+1:t+H}$ denote an action chunk and $d^0=d_{t+1:t+H}$ the corresponding future DINO features. Gaussian noise and flow times are sampled independently for the two streams:
\begin{equation}
a^{\tau_a}=(1-\tau_a)a^0+\tau_a\epsilon_a,\qquad
d^{\tau_d}=(1-\tau_d)d^0+\tau_d\epsilon_d,
\end{equation}
where $\epsilon_a,\epsilon_d\sim\mathcal N(0,I)$ and $\tau_a$ and $\tau_d$ are sampled independently. Conditioned on the current visual features $d_t$, robot state $s_t$, and instruction embedding $l$, the model predicts an action velocity and a future-feature velocity:
\begin{equation}
(v_\theta^a,v_\theta^d)=
f_\theta(a^{\tau_a},d^{\tau_d},d_t,s_t,l,\tau_a,\tau_d).
\end{equation}
Both predictions are optimized in the same forward pass, with the future-feature objective weighted by $\lambda_{\mathrm{world}}$:
\begin{equation}
\mathcal L =
\mathbb E\!\left[ \|v_\theta^a-(\epsilon_a-a^0)\|_2^2+ \lambda_{\mathrm{world}} \|v_\theta^d-(\epsilon_d-d^0)\|_2^2 \right].
\end{equation}

The two prediction streams are separated by an attention mask as shown in Figure~\ref{fig:framework}. Context tokens attend within the current context; action tokens attend to the context and action stream; and future-feature tokens attend to the context and world stream. Thus, action tokens cannot access future-feature tokens during training.

At inference, \name omits the future-feature tokens and iteratively denoises only the action chunk. Future DINO features are training targets, not inputs or generated outputs of the deployed policy. This separates the benefit of future-feature supervision from the cost of generating a future at test time.

\vspace{0.2cm}
\noindent
\textbf{Architecture.} $f_\theta$ is parameterized with a Mixture-of-Transformers backbone~\cite{liang2024mixture} containing modality-specific context, action, and world experts~\cite{esser2024scaling}. Current DINO features, language, and proprioception form the context stream, while actions and future DINO features form separate generative streams. Each expert uses modality-specific projections and feed-forward layers, while the structured attention mask controls information flow between streams. Separate output heads predict the action and future-feature velocity fields.

\vspace{0.2cm}
\noindent
\textbf{Inference.} At inference time, only the context and the action streams are instantiated. Future-feature tokens and the world prediction head are removed, and \name directly denoises the action chunk conditioned on the current visual observations, language instruction, and proprioceptive state. Consequently, future-feature prediction introduces no additional future-generation cost at deployment.

\section{Results and Discussion} \label{sec:evaluation}
We evaluate how the design of future-feature supervision affects world-action learning in pretrained visual representation spaces. Specifically, we address four questions: (1) Does future-feature prediction improve policy learning over action-only training? (2) How does the choice of visual representation affect performance? (3) How sensitive is policy learning to the design of future supervision, including target viewpoint, feature parameterization, and temporal density? (4) Is future prediction more effective as auxiliary supervision or through direct action-world interaction?

\subsection{Evaluation in Simulation Environment}
We evaluate \name on two multi-task simulation benchmarks: LIBERO~\cite{liu2023libero} and LIBERO-Plus~\cite{fei2025libero}. For LIBERO, we use the official datasets and report success rates across four suites: Spatial, Object, Goal, and Long with 50 rollouts per task. LIBERO-Plus extends the original LIBERO tasks with perturbations that evaluate robustness under conditions not observed during standard training. We evaluate the LIBERO-trained policies directly on LIBERO-Plus without additional training.

\subsection{Results}
\textbf{LIBERO.} Table~\ref{tab:libero} compares \name with representative VLA and WAM baselines across the four LIBERO suites. Despite using only 0.4B parameters and no embodied pretraining, \name achieves a 97.6\% average success rate, matching the substantially larger Fast-WAM (97.6\%) and outperforming Light-WAM (97.2\%). \name also remains competitive with VLAs that benefit from large-scale embodied pretraining, including OpenVLA-OFT (97.1\%) and $\pi_{0.5}$ (96.9\%), while using considerably fewer parameters.

\vspace{0.2cm}
\noindent
\textbf{LIBERO-Plus.} Table~\ref{tab:liberoplus} evaluates zero-shot robustness under seven perturbation types in LIBERO-Plus. Despite using only 0.4B parameters and no embodied pretraining, \name achieves a 74.4\% average success rate, outperforming all other non-pretrained WAM baselines, including Fast-WAM (49.5\%), OpenWAM-$\alpha$ (69.2\%), Faster-WAM (73.6\%), and BridgeWAM (72.2\%). \name also achieves the highest performance under camera perturbations (87.3\%) and remains competitive under background, noise, and layout shifts.

\subsection{Ablation Studies}
To identify which design choices shape \name's performance, we study action-world coupling, target parameterization, temporal density, future-target viewpoint, visual encoder choice, and visual context representation. We additionally evaluate these ablations under zero-shot LIBERO-Plus perturbations and provide further analyses of future-target horizons, loss weighting, token budget, and alternative coupling designs in the supplementary material.

\begin{figure}[htbp]
    \includegraphics[width=1.0\linewidth]{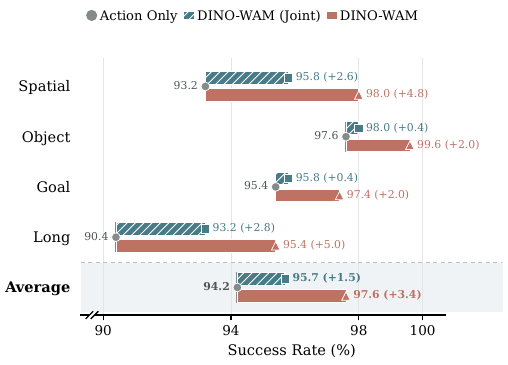}
    \centering
    \captionof{figure}{\small \textbf{Effect of Action-World Attention.} LIBERO success of the action-only baseline, masked \name, and \name (Joint). The masked formulation achieves the highest performance, showing that future-feature prediction is most beneficial as an auxiliary training signal.}
    \label{fig:future-supervision}
\end{figure}

\vspace{0.2cm}
\noindent
\textbf{Effect of Action-World Attention.} We further study whether direct interaction between the action and future-feature streams is beneficial by training an unmasked joint-generation variant, \name (Joint), with the same flow-matching objective. Removing the structured attention mask allows action and future-feature tokens to interact, and both streams are generated at inference. As shown in Figure~\ref{fig:future-supervision}, the action-only baseline achieves an average LIBERO success rate of 94.2\%, compared with 97.6\% for the masked \name and 95.7\% for \name (Joint). These results suggest that future-feature prediction is most effective when used as an auxiliary training signal, while direct action-world coupling through joint generation provides a smaller improvement.

\begin{figure}[htbp]
    \includegraphics[width=0.95\linewidth]{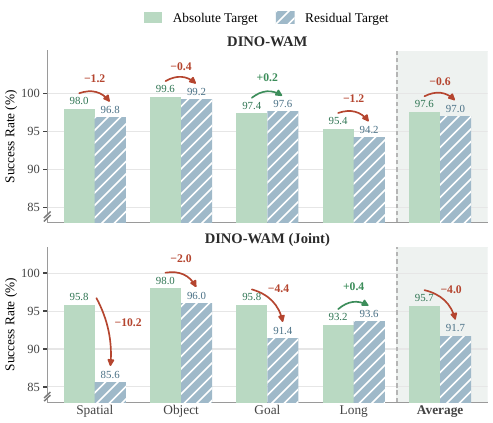}
    \centering
    \captionof{figure}{\small \textbf{Absolute vs. Residual Feature Supervision.} LIBERO success rate of predicting future DINO features directly versus predicting temporal feature residuals for \name and \name (Joint). Absolute targets achieve higher average success in both settings.}
    \label{fig:absolutevsresidual}
\end{figure}

\vspace{0.2cm}
\noindent
\textbf{Absolute vs. Residual Feature Supervision.} Figure~\ref{fig:absolutevsresidual} compares direct prediction of future DINO features with prediction of their temporal residuals. Absolute targets yield a higher four-suite average for both \name (97.6\% versus 97.0\%) and \name (Joint) (95.7\% versus 91.7\%). Notably, residual supervision substantially degrades the joint-generation variant, reducing its performance below the action-only baseline. These results indicate that absolute DINO features provide a more stable and effective target for future-feature prediction, particularly under direct action--world coupling.

\begin{figure}[htbp]
    \centering
    \includegraphics[width=0.95\linewidth]{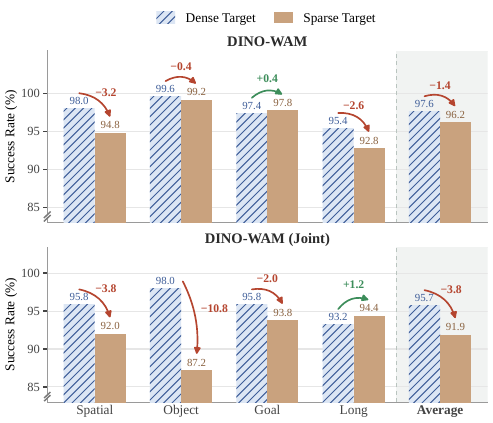}
    \captionof{figure}{\small \textbf{Dense vs. Sparse Feature Supervision.} LIBERO success rates for dense and temporally sparse future-feature targets for \name and \name (Joint). Dense supervision achieves higher average performance in both settings.}
    \label{fig:densevssparse}
\end{figure}

\vspace{0.2cm}
\noindent
\textbf{Dense vs. Sparse Feature Supervision.} Figure~\ref{fig:densevssparse} compares dense near-term future supervision with temporally sparse targets extending beyond the action horizon. Dense supervision achieves a higher four-suite average for both \name (97.6\% versus 96.2\%) and \name (Joint) (95.7\% versus 91.9\%). Notably, sparse supervision substantially degrades the joint-generation variant, reducing its performance below the action-only baseline. These results suggest that, in our setting, dense near-term future-feature supervision provides a stronger and more reliable learning signal than sparsely distributed long-horizon targets.

\begin{table}[htbp]
\centering
\captionof{table}{\small \textbf{Effect of Future-Target Camera View.} LIBERO success rates for \name and \name (Joint) using wrist, agent, or dual-camera future-feature supervision while keeping the current visual context fixed to both cameras. Best results are \textbf{bold}, and second-best results are \underline{underlined}.}
\label{tab:camera_view}
\resizebox{0.45\textwidth}{!}{%
\begin{tabular}{lc|cccc|c}
\toprule
\rowcolor{gray!20} \textbf{Model} & \textbf{View} & \textbf{Spatial} $\uparrow$ & \textbf{Object} $\uparrow$ & \textbf{Goal} $\uparrow$& \textbf{Long} $\uparrow$& \textbf{Average} $\uparrow$\\
\midrule
\small \textbf{Action Only}&\small -&93.2\%&97.6\%&95.4\%&90.4\%&94.2\% \\
\midrule
\multirow{3}{*}{\small \textbf{\name}}
 & \small \textbf{Wrist}      & \textbf{98.0}\% & 98.6\% & 96.4\% & 94.6\% & \underline{96.9}\% \\
 & \small \textbf{Agent}      & 95.2\% & \underline{99.2}\% & 95.8\% & 95.0\% & 96.3\% \\
 & \small \textbf{Both}       & \textbf{98.0}\% & \textbf{99.6}\% & \textbf{97.4}\% & \underline{95.4}\% & \textbf{97.6}\% \\
\midrule
\multirow{3}{*}{\small \shortstack{\textbf{\name} \\ \textbf{Joint}}}
 & \small \textbf{Wrist}      & 95.2\%   & 98.8\%   & \textbf{97.4}\%   & \textbf{95.8}\%   & 96.8\% \\
 & \small \textbf{Agent}      & 94.4\%   & 97.4\%   & \underline{97.2}\%   & 94.8\%   & 96.0\% \\
 & \small \textbf{Both}       & \underline{95.8}\% & 98.0\% & 95.8\% & 93.2\% & 95.7\% \\
\bottomrule
\end{tabular}}
\end{table}

\vspace{0.2cm}
\noindent
\textbf{Future-Target Camera View.} Table~\ref{tab:camera_view} studies how the source of future-feature supervision affects performance while keeping the current two-camera context unchanged. For \name, using both future views achieves the highest average success rate (97.6\%), compared with wrist-only (96.9\%) and agent-only (96.3\%) supervision. In contrast, \name (Joint) performs best with wrist-only targets (96.8\%), while dual-view supervision yields a lower average of 95.7\%. Overall, these results suggest that the most effective future-view target depends on how the world stream is used: multi-view supervision performs best for the masked auxiliary formulation, whereas joint action--world generation favors the more local wrist-view representation.

\begin{table}[htbp]
\centering
\captionof{table}{\small \textbf{Effect of Visual Encoder Choice.} LIBERO success rates (\%) for \name and \name (Joint) using DINOv2~\cite{oquab2023dinov2}, SigLIP~\cite{zhai2023sigmoid}, and V-JEPA 2~\cite{assran2025v} encoders for both current visual context and future feature supervision. Best results are \textbf{bold}, and second-best results are \underline{underlined}.}
\label{tab:encoder_choice}
\resizebox{0.49\textwidth}{!}{%
\begin{tabular}{l|lc|cccc|c}
\toprule
\rowcolor{gray!20} \textbf{Model} & \textbf{Encoder} & \textbf{Params} & \textbf{Spatial} $\uparrow$ & \textbf{Object} $\uparrow$ & \textbf{Goal} $\uparrow$ & \textbf{Long} $\uparrow$ & \textbf{Average} $\uparrow$ \\
\midrule
\multirow{3}{*}{\small \textbf{\name}}
 & \small \textbf{DINOv2}~\cite{oquab2023dinov2} &  21M& \textbf{98.0}\% & \textbf{99.6}\% & \textbf{97.4}\% & \textbf{95.4}\% & \textbf{97.6}\% \\
 & \small \textbf{SigLIP}~\cite{zhai2023sigmoid} &  92.9M& 93.0\% & 98.0\% & 95.4\% & 93.6\% & 95.0\% \\
 & \small \textbf{V-JEPA 2}~\cite{assran2025v} & 304M& 92.8\%   & \underline{98.6}\%   & 94.8\%   & 89.0\%   & 93.8\% \\
\midrule
\multirow{3}{*}{\small \shortstack{\textbf{\name} \\ \textbf{Joint}}}
 & \small \textbf{DINOv2}~\cite{oquab2023dinov2} &  21M& \underline{95.8}\% & 98.0\% & \underline{95.8}\% & 93.2\% & 95.7\% \\
 & \small \textbf{SigLIP}~\cite{zhai2023sigmoid} & 92.9M& 91.2\% & \textbf{99.6}\% & \textbf{97.4}\% & \underline{95.0}\% & \underline{95.8}\% \\
 & \small \textbf{V-JEPA 2}~\cite{assran2025v} & 304M& 91.2\%   & 96.6\%   & 92.6\%   & 91.2\%   & 92.9\% \\
\bottomrule
\end{tabular}}
\end{table}

\vspace{0.2cm}
\noindent
\textbf{Choice of Visual Encoder.} Table~\ref{tab:encoder_choice} studies how the pretrained visual representation affects future-feature supervision by replacing DINOv2 with SigLIP and V-JEPA 2 while keeping the rest of the training setup fixed. Implementation details for adapting each encoder to the shared \name architecture are provided in the supplementary material. For the masked \name formulation, DINOv2 performs best across all four LIBERO suites, achieving an average success rate of 97.6\%, compared with 95.0\% for SigLIP and 93.8\% for V-JEPA 2. Under joint generation, the gap narrows substantially: SigLIP achieves the highest average (95.8\%), closely followed by DINOv2 (95.7\%), while V-JEPA 2 remains lower at 92.9\%. These results indicate that DINOv2 provides the strongest visual representation among the evaluated encoders for the masked formulation.

\begin{figure}[htbp]
    \centering
    \includegraphics[width=0.95\linewidth]{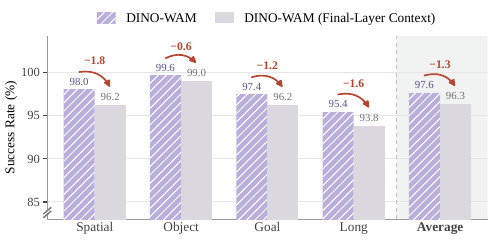}
    \captionof{figure}{\small \textbf{Effect of DINO Context Layers.} LIBERO success rates using the final DINOv2 block alone versus fused multi-layer context features. Multi-layer fusion improves performance across all task suites.}
    \label{fig:context_layers}
\end{figure}

\vspace{0.2cm}
\noindent
\textbf{Effect of DINO Context Layers.} Figure~\ref{fig:context_layers} compares using only the final DINOv2 block with fusing features from multiple intermediate layers for the current visual context, while keeping the future-feature targets fixed. Multi-layer fusion improves performance across all LIBERO suites, increasing the average success rate from 96.3\% to 97.6\%. These results indicate that intermediate DINO features provide complementary information beyond the final-layer representation and improve policy learning.

\subsection{Real-world Deployment}
We further evaluate \name on contact-rich real-world disassembly tasks.

\begin{figure}[htbp]
    \includegraphics[width=0.9\linewidth]{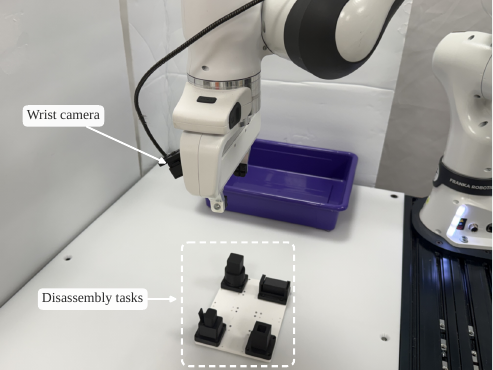}
    \centering
    \captionof{figure}{\small \textbf{Real-World Deployment Setup.} We evaluate \name on real-world disassembly tasks using a 7-DoF Franka robot equipped with two RGB cameras for visual observation.}
    \label{fig:setup}
\end{figure}

\begin{figure*}[t]
    \includegraphics[width=1.0\linewidth]{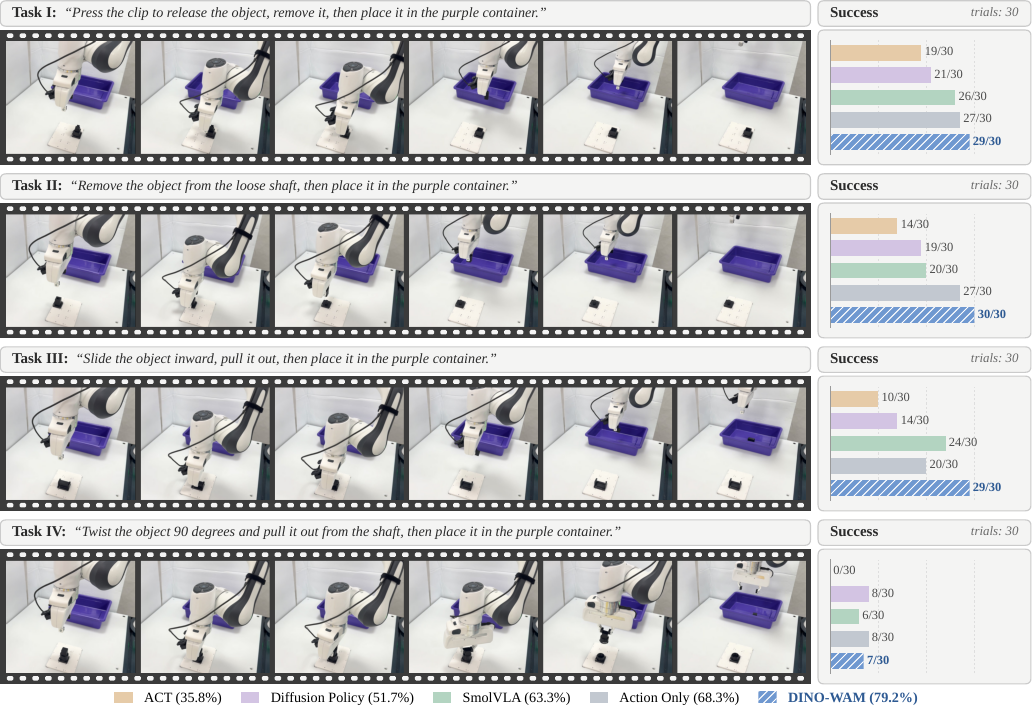}
    \centering
    \captionof{figure}{\small \textbf{Real-world Deployment of \name.} Examples of \name executing four real-world disassembly tasks, together with success-rate comparisons against baseline policies.}
    \label{fig:realdworld_deployment}
\end{figure*}

\vspace{0.2cm}
\noindent
\textbf{Experimental Setup \& Tasks.} Our real-world platform consists of a 7-DoF Franka Emika Panda manipulator equipped with a parallel-jaw gripper. Visual observations are captured by two RGB cameras: a fixed third-person camera that provides a global view of the workspace and a wrist-mounted camera that captures close-range, fine-grained observations during manipulation, as illustrated in Figure~\ref{fig:setup}.

We evaluate \name on a set of contact-rich real-world disassembly tasks that require precise visual reasoning and fine-grained manipulation as shown in Figure~\ref{fig:setup}. The task suite and demonstration dataset follow the setup introduced in~\cite{soleymanzadeh2026phr}. Each task is specified by a natural-language instruction and contains 100 human-teleoperated demonstrations recorded at 10~Hz. Each trajectory includes synchronized RGB observations from a fixed third-person camera and a wrist-mounted camera, robot proprioceptive states, action commands, and the corresponding language instruction.

\vspace{0.2cm}
\noindent
\textbf{Results \& Analysis.} Figure~\ref{fig:realdworld_deployment} reports success rates across four real-world disassembly tasks and illustrates representative successful rollouts of \name. \name achieves the highest average success rate at 79.2\% (95/120), outperforming ACT~\cite{zhao2023learning} (35.8\%), Diffusion Policy~\cite{chi2025diffusion} (51.7\%), SmolVLA~\cite{shukor2025smolvla} (63.3\%), and the matched action-only baseline (68.3\%). The 10.9-point improvement over the matched action-only baseline provides direct evidence that future-feature prediction supplies a useful auxiliary learning signal for contact-rich manipulation.

\section{Conclusions} \label{sec:conclusions}
We introduced \name, a compact world-action model that augments robot policy learning with future prediction in a pretrained visual feature space. Rather than reconstructing future RGB observations, \name predicts future visual features and uses an attention mask to keep the world stream auxiliary to action generation. Across LIBERO and LIBERO-Plus, \name remains competitive with substantially larger VLA and WAM models despite using only 0.4B parameters and no embodied pretraining, and it further improves over an action-only counterpart on real-world contact-rich disassembly tasks. Our ablations show that the design of feature-space world modeling matters: masked auxiliary prediction outperforms direct joint generation, while DINOv2 features, absolute targets, dense near-term supervision, multi-view future targets, and multi-layer visual context perform best in our primary LIBERO setting.

\vspace{0.2cm}
\noindent
\textbf{Limitations.} Our study is primarily evaluated on LIBERO and LIBERO-Plus, with additional validation on a small set of real-world disassembly tasks. While these benchmarks provide controlled settings for analyzing feature-space world-action learning, they do not capture the full diversity of robotic manipulation scenarios, embodiments, and interaction dynamics. Our experiments also consider a limited set of pretrained visual encoders and model scales, and all visual encoders remain frozen during policy training. Future work should evaluate whether the observed design trends persist across larger datasets, different robot embodiments, broader task distributions, and jointly adapted visual representations.

\section*{Acknowledgments}
This work was supported by the U.S. National Science Foundation under Grant Nos. 2422826 and 2527316. Portions of this research were conducted using the advanced computing resources provided by Texas A\&M High Performance Research Computing (HPRC).

\section*{LLM Usage Statement}
We used Claude and Codex to support idea development, method discussion, and refinement of the overall research direction. ChatGPT was used for language polishing and stylistic refinement, as well as for suggestions during figure preparation and visualization design. All technical decisions, experiments, results, and final manuscript content were reviewed and verified by the authors.

{
    \small
    \bibliographystyle{ieeenat_fullname}
    \bibliography{main}
}

\clearpage

\maketitlesupplementary
We first provide additional implementation details for \name, including its architecture, hyperparameters, training setup, and visual feature construction. Then we describe the architecture-matched action-only baseline. We subsequently present additional analyses of future-feature supervision, including target design, action-world coupling, and alternative supervision architectures, followed by multi-seed evaluation, robustness under distribution shift, and inference-efficiency results.

\section{Implementation Details} \label{app:structure}
\subsection{Structure \& Hyperparameters}
\name consists of frozen DINOv2~\cite{oquab2023dinov2} and CLIP~\cite{radford2021learning} encoders followed by a Mixture-of-Transformers~\cite{liang2024mixture} backbone with three modality-specific experts: context, action, and world. Each expert has its own normalization, QKV projections, output projections, and feed-forward layers, while attention is computed jointly under the attention mask described in the main paper. The current visual context is constructed from multiple DINOv2 blocks, whereas future-feature targets are extracted from the final DINOv2 block. \Cref{tab:hparams} summarizes the full model and training configuration.

\begin{table}[htbp]
\centering
\small
\setlength{\tabcolsep}{6pt}
\captionof{table}{\small \textbf{\name hyperparameters.} Main architecture and training configuration used in our experiments.}
\label{tab:hparams}
\resizebox{0.9\linewidth}{!}{%
\begin{tabular}{@{}ll@{}}
\toprule
\rowcolor{gray!20} \textbf{Component} & \textbf{Value} \\
\midrule
\multicolumn{2}{@{}l}{\emph{Encoders (frozen)}} \\
\quad Visual encoder & DINOv2 ViT-S/14~\cite{oquab2023dinov2} \\
\quad Input resolution & $224\times224$ \\
\quad Context blocks & $\{2,5,8,11\}$ \\
\quad Future-target block & $11$ \\
\quad Tokens per camera & $16$ \\
\quad Language encoder & CLIP ViT-B/32~\cite{radford2021learning} \\
\midrule
\multicolumn{2}{@{}l}{\emph{Backbone}} \\
\quad Architecture & Mixture-of-Transformers~\cite{liang2024mixture} \\
\quad Experts & context / action / world \\
\quad Transformer layers & $16$ \\
\quad Hidden dimension & $768$ \\
\quad Attention heads & $12$ \\
\quad MLP ratio & $4$ \\
\quad Training parameters & $346.1$M \\
\midrule
\multicolumn{2}{@{}l}{\emph{Prediction}} \\
\quad Action horizon $H$ & $10$ \\
\quad Future-target horizons & $\{1,2,\dots,10\}$ \\
\quad Future-feature target & absolute DINOv2 features \\
\quad Future-feature loss weight $\lambda_{\mathrm{world}}$ & $1.0$ \\
\midrule
\multicolumn{2}{@{}l}{\emph{Flow matching}} \\
\quad Flow-time distribution & $\mathcal{U}(0,1)$ \\
\quad Sampling steps & $20$ \\
\quad ODE solver & Euler \\
\bottomrule
\end{tabular}}
\end{table}

\subsection{Training Details}
We optimize all models using AdamW with a learning rate of $1\times10^{-4}$, weight decay of $10^{-6}$, and a cosine learning-rate schedule with 2,000 warm-up steps. Models are trained for 40k optimization steps with an effective batch size of 128. Training uses mixed precision and gradient clipping with a maximum norm of 1.0. Unless otherwise specified, all experiments are trained on a single NVIDIA A100 40GB GPU.

\subsection{Visual Feature Extraction}
\subsubsection{DINOv2}
All visual features are extracted using a frozen DINOv2 ViT-S/14~\cite{oquab2023dinov2}. Given an input resolution of $224\times224$ and a patch size of 14, DINOv2 produces a $16\times16$ grid of patch tokens. We denote the layer-normalized patch features from block $\ell$ at timestep $t$ as
$P^{\ell}_t\in\mathbb{R}^{16\times16\times C}$, where $C=384$, and discard the class token.

\vspace{0.2cm}
\noindent
\textbf{Spatial Pooling.} To reduce the visual token budget, each feature grid is average-pooled over non-overlapping $4\times4$ spatial windows, producing $N=16$ tokens per camera:
\begin{equation}
\bar{p}^{\,\ell}_{t,(i,j)} = \frac{1}{16} \sum_{u=0}^{3}\sum_{v=0}^{3} P^{\ell}_{t,(4i+u,\,4j+v)}, \qquad i,j\in\{0,\dots,3\}.
\end{equation}

\vspace{0.2cm}
\noindent
\textbf{Multi-layer Context.} For the current visual observation, the main \name configuration uses features from DINOv2 blocks $\mathcal{B}=\{2,5,8,11\}$. For each spatial token $k$, the pooled features from these blocks are concatenated and fused through a learned linear projection:
\begin{equation}
d_{t,k} = W \big[\bar{p}^{\,2}_{t,k};\bar{p}^{\,5}_{t,k}; \bar{p}^{\,8}_{t,k}; \bar{p}^{\,11}_{t,k} \big] +b, \qquad W\in\mathbb{R}^{C\times4C}.
\end{equation}
We initialize $W=[\,0\;0\;0\;I\,]$ and $b=0$, such that the fused representation initially corresponds to the final-layer features and can subsequently learn to incorporate information from intermediate layers. Tokens from all camera views are concatenated to form the visual context $d_t\in\mathbb{R}^{16V\times C}$, where $V$ denotes the number of cameras.

\vspace{0.2cm}
\noindent
\textbf{Future-Feature Targets.} Future-feature targets are extracted from the final DINOv2 block. For a future horizon $h$, we define $d_{t+h}=\bar{p}^{\,11}_{t+h}$. Depending on the target parameterization, the world stream predicts either the absolute future feature or its temporal residual relative to the current observation:
\begin{equation}
x_h=d_{t+h} \quad\text{(absolute)},\qquad x_h=d_{t+h}-\bar{p}^{\,11}_{t} \quad\text{(residual)}.
\end{equation}
The main \name configuration uses absolute future-feature targets.

\begin{figure*}[t]
  \centering
  \includegraphics[width=0.8\linewidth]{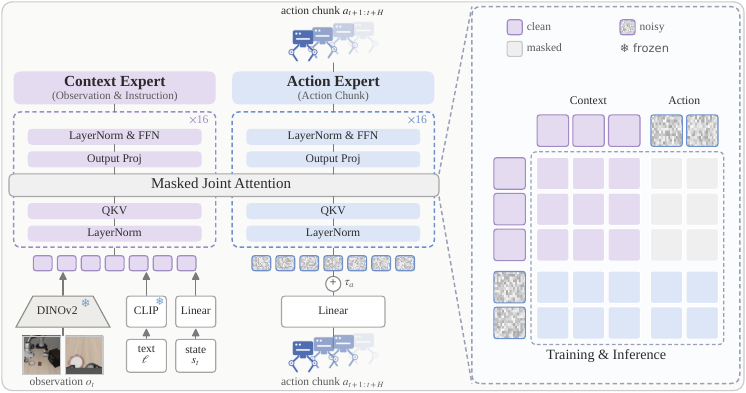}
  \caption{\small \textbf{Action-Only Objective Baseline.} \textit{Left}: The baseline retains \name's context and action pathways. Frozen DINOv2~\cite{oquab2023dinov2} features from the current camera views, a frozen CLIP~\cite{radford2021learning} instruction embedding, and the robot state form the context tokens. The action expert denoises a 10-step action chunk using flow matching. Training uses the same masked Mixture-of-Transformers~\cite{liang2024mixture} architecture as \name, but optimizes only the action loss. \textit{Right}: Context tokens attend only to context; action tokens attend to context and other action tokens. At inference, only the context and action pathways are evaluated, with 20 Euler steps to generate the action chunk.}
  \label{fig:supp_actiononly}
\end{figure*}

\subsubsection{Adapting SigLIP and V-JEPA 2}
For the visual-encoder ablation, we replace the frozen DINOv2 image encoder with either SigLIP-Base (patch size 16, feature dimension 768)~\cite{zhai2023sigmoid} or V-JEPA 2 ViT-L (feature dimension 1024)~\cite{assran2025v}. The CLIP text encoder, two camera views, robot-state input, action chunk, and training objectives remain unchanged.

For SigLIP, we extract patch features from transformer blocks $\{2,5,8,11\}$ and apply the model's post-layer normalization to each selected block. For V-JEPA 2, each image is resized to $256\times256$. Because its patch embedding uses a two-frame tubelet, we repeat the same image twice to form a single tubelet. Consequently, each extracted representation corresponds to a single observation. We extract patch features from blocks $\{5,11,17,23\}$ and apply layer normalization to the resulting tokens.

For both encoders, patch features from each selected block are spatially average-pooled to $16$ tokens per camera. At each spatial location, features from the four selected blocks are concatenated along the feature dimension and passed through an encoder-specific learned fusion projection, following the same multi-layer fusion design used for DINOv2. Future-feature targets are constructed from the encoder's final-layer features and similarly pooled to $16$ tokens per camera at each of the future horizons. The downstream model retains the same 16-layer, 768-dimensional Mixture-of-Transformers backbone, with encoder-specific input and target projections accommodating the native feature dimension of each visual encoder. We train both using the same respective objectives and training protocol as their DINOv2 counterparts.

\section{Action-Only Baseline} \label{app:action-only}
To isolate the contribution of future-feature supervision, we train an architecture-matched action-only counterpart of \name as shown in Figure~\ref{fig:supp_actiononly}. It uses the same current DINO features, proprioception, language input, context and action experts, and attention mask as the main model.

For a clean action chunk $a^0=a_{t+1:t+H}$, we sample $\epsilon\sim\mathcal{N}(0,I)$ and construct
\begin{equation}
a^\tau=(1-\tau)a^0+\tau\epsilon.
\end{equation}
The action pathway predicts its velocity from the noisy action and current
context,
\begin{equation}
v_\theta^a=f_\theta^a(a^\tau,d_t,s_t,l,\tau),
\end{equation}
and is optimized with
\begin{equation}
\mathcal{L}_{\mathrm{action\text{-}only}}
=\mathbb{E}\!\left[
\left\|v_\theta^a-(\epsilon-a^0)\right\|_2^2
\right].
\end{equation}
This ablation variant retains the main model's action architecture and training protocol while removing the world stream.

\vspace{0.2cm}
\noindent
\textbf{Architecture.} The action-only-objective baseline isolates the effect of future-feature supervision within the architecture of \name. It uses the same frozen DINOv2 and CLIP encoders, learned fusion of DINO blocks $\{2,5,8,11\}$, language and state tokens, 10-step action chunk, and masked context and action experts. The context and action pathway has 228.8M parameters. Training uses an effective batch size of 128 and otherwise follows the settings above.

\vspace{0.2cm}
\noindent
\textbf{Inference.} Only context and action tokens are processed at inference. Actions are generated from Gaussian noise using 20 Euler steps, exactly as in \name's masked-policy inference.

\section{Additional Analysis of Future Supervision}
\subsection{Future-Target Design}
\begin{figure}[t]
    \centering
    \includegraphics[width=0.95\linewidth]{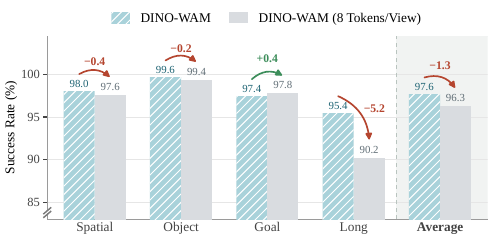}
    \caption{\small \textbf{Future-feature Token Budget.} LIBERO success rates (\%) with 16 or 8 future DINO tokens per camera at each of the 10 target horizons. Current-observation tokens and all other settings are unchanged.}
    \label{fig:future_token_budget}
\end{figure}

\noindent
\textbf{Token Budget.} Figure~\ref{fig:future_token_budget} tests the spatial granularity of future supervision while keeping the current visual context fixed. Reducing the future targets from 16 to 8 tokens per camera lowers the four-suite average from 97.6\% to 96.3\%. Most of this difference comes from LIBERO-Long, where success falls from 95.4\% to 90.2\%.

\vspace{0.2cm}
\noindent
\textbf{Future-Target Horizon.} We further study how the temporal placement and density of future-feature targets affect policy learning. The main \name configuration uses dense supervision over the next ten timesteps, matching the action horizon. We compare this setting with single-step targets at $\{10\}$ and $\{30\}$, a sparse near-term schedule $\{3,6,9\}$, a sparse long-range schedule $\{3,15,30\}$, and a denser long-horizon schedule $\{3,6,9,\ldots,30\}$. As shown in Table~\ref{tab:future_horizon}, all configurations achieve relatively similar performance, with average LIBERO success ranging from 96.2\% to 97.6\%. Dense near-term supervision achieves 97.6\%, while a single target at timestep 10 and the sparse $\{3,15,30\}$ schedule both reach 97.2\%. Extending supervision to more distant horizons does not provide a consistent improvement: the single target at timestep 30 and the denser long-horizon schedule $\{3,6,9,\ldots,30\}$ both achieve 96.2\%.

\begin{table}[htbp]
\centering
\captionof{table}{\small \textbf{Effect of Future-Target Horizon.} \name's LIBERO success rates (\%) for different temporal placements of future-feature targets. Best results are \textbf{bold}, and second-best results are \underline{underlined}.}
\label{tab:future_horizon}
\resizebox{0.95\linewidth}{!}{
\begin{tabular}{l|cccc|c}
\toprule
\rowcolor{gray!20} \small \textbf{Future Targets} &\small \textbf{Spatial} $\uparrow$ &\small \textbf{Object} $\uparrow$ &\small \textbf{Goal} $\uparrow$ &\small \textbf{Long} $\uparrow$ & \small \textbf{Average} $\uparrow$ \\
\midrule
$\{10\}$&\small 97.4\%&\small \textbf{99.6}\%&\small 96.4\%&\small \underline{95.4}\%&\small \underline{97.2}\% \\
$\{3,6,9\}$ &\small 97.0\% &\small \underline{99.2}\% &\small 96.4\% &\small 93.4\% &\small 96.5\% \\
$\{1, 2, 3, \cdots, 10\}$ &\small \underline{98.0}\% &\small \textbf{99.6}\% &\small \underline{97.4}\% &\small \underline{95.4}\% &\small \textbf{97.6}\% \\
$\{30\}$&\small 96.0\%&\small 98.8\%&\small 95.8\%&\small 94.2\%&\small 96.2\% \\
$\{3,15,30\}$ &\small \textbf{98.4}\% &\small 98.0\% &\small 96.4\% &\small \textbf{95.8}\% &\small \underline{97.2}\% \\
$\{3, 6, 9, \cdots, 30\}$&\small 94.8\% &\small \underline{99.2}\% &\small \textbf{97.8}\% &\small 92.8\% &\small 96.2\% \\
\bottomrule
\end{tabular}}
\end{table}

\begin{figure*}[t]
    \centering
    \includegraphics[width=0.95\linewidth]{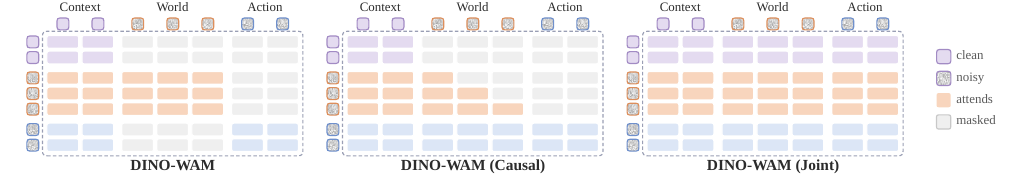}
    \captionof{figure}{\small \textbf{Action--World Attention Patterns.} Comparison of the attention structures used in the three \name coupling variants. \emph{Masked} keeps the action and world streams separated: both attend to the shared context and within their own streams, but not to one another. \emph{Causal} introduces directional world-to-action coupling, allowing action tokens to attend to predicted future-feature tokens while preventing the reverse interaction. \emph{Joint} enables bidirectional attention between the action and world streams, allowing both modalities to directly influence one another during generation. These patterns isolate how the degree and direction of action--world information flow affect policy learning.}
    \label{fig:attention_pattern}
\end{figure*}

\vspace{0.2cm}
\noindent
\textbf{Loss Weight.} We also examine the sensitivity of \name to the relative weight assigned to the future-feature prediction objective. Keeping the action loss unchanged, we vary the future-feature loss weight $\lambda_{\mathrm{world}}$. The default setting $\lambda_{\mathrm{world}}=1$ achieves the highest average LIBERO success rate of \underline{97.6}\%, while weights of 0.25 and 2.0 achieve \underline{97.3}\% and \underline{97.2}\%, respectively. Thus, performance remains relatively stable across a broad range of nonzero weights.

\begin{table}[htbp]
\centering
\caption{\textbf{Effect of Future-Feature Loss Weight.}
LIBERO success rates (\%) when varying the weight $\lambda_{\mathrm{world}}$ of the future-feature prediction objective. Best results are \textbf{bold}, and second-best results are \underline{underlined}.}
\label{tab:world_loss_weight}
\resizebox{0.95\linewidth}{!}{
\begin{tabular}{l|cccc|c}
\toprule
\rowcolor{gray!20} $\lambda_{\mathrm{world}}$ &\small \textbf{Spatial} $\uparrow$ &\small \textbf{Object} $\uparrow$ &\small \textbf{Goal} $\uparrow$ &\small \textbf{Long} $\uparrow$ &\small \textbf{Average} $\uparrow$ \\
\midrule
\small 0.25 &\small 96.4\% &\small \textbf{99.6}\% &\small \textbf{98.0}\% &\small \underline{95.2}\% &\small \underline{97.3}\% \\
\small \textbf{1.0 (main)} &\small \textbf{98.0}\% &\small \textbf{99.6}\% &\small \underline{97.4}\% &\small \textbf{95.4}\% &\small \textbf{97.6}\% \\
\small 2.0 &\small \underline{97.8}\% &\small \underline{98.8}\% &\small \textbf{98.0}\% &\small 94.4\% &\small 97.2\% \\
\bottomrule
\end{tabular}}
\end{table}

\vspace{0.2cm}
\noindent
\textbf{Context Feature Fusion.} We further examine how features from different DINOv2 layers should be combined to construct the current visual context. Our main configuration learns a fusion of features from blocks 2, 5, 8, and 11. We compare this with simple mean fusion of the same four blocks and with using only the final DINOv2 layer (11). As shown in Table~\ref{tab:context_fusion}, learned fusion achieves the highest performance across all four LIBERO suites, with an average success rate of \underline{97.6}\%. Mean fusion remains competitive at \underline{97.0}\%, while using only the final layer reduces the average to \underline{96.3}\%. These results indicate that intermediate DINO features provide complementary information beyond the final-layer representation.

\begin{table}[htbp]
\centering
\captionof{table}{\textbf{Effect of DINO Context Feature Fusion.} \name LIBERO success rates (\%) using different strategies for constructing the current visual context from DINOv2 features. Best results are \textbf{bold}, and second-best results are \underline{underlined}.}
\label{tab:context_fusion}
\resizebox{0.95\linewidth}{!}{
\begin{tabular}{l|cccc|c}
\toprule
\rowcolor{gray!20} \small \textbf{Context Features} & \small \textbf{Spatial} $\uparrow$ & \small \textbf{Object} $\uparrow$ & \small \textbf{Goal} $\uparrow$ & \small \textbf{Long} $\uparrow$ & \small \textbf{Average} $\uparrow$ \\
\midrule
\small \textbf{Learned Fusion} & \small \textbf{98.0}\% & \small \textbf{99.6}\% & \small \textbf{97.4}\% & \small \textbf{95.4}\% & \small \textbf{97.6}\% \\
\midrule
\small \textbf{Mean Fusion} & \small \underline{97.8}\% & \small \underline{99.4}\% & \small \underline{96.6}\% & \small \underline{94.4}\% & \small \underline{97.0}\% \\
\midrule
\small \textbf{Final Layer} & \small 96.2\% & \small 99.0\% & \small 96.2\% & \small 93.8\% & \small 96.3\% \\
\bottomrule
\end{tabular}}
\end{table}

\subsection{Action-World Coupling}
We further study how information flow between the world and action streams affects policy learning. In the default \name formulation, the action and world streams independently attend to the shared context but cannot attend to one another, keeping future prediction auxiliary to action generation. We compare this design with two alternatives: \name (Causal), where the world stream attends to context and predicted future features while the action stream additionally attends to the world stream, allowing future predictions to directly guide action generation; and \name (Joint), where action and world tokens attend bidirectionally to one another and are generated together as shown in Figure~\ref{fig:attention_pattern}.

As shown in Table~\ref{tab:coupling_camera_view}, all three causal variants substantially outperform the action-only baseline and the bidirectionally coupled joint model. The wrist- and agent-view causal variants both achieve 97.4\% average success, while the dual-camera causal variant reaches 97.1\%. However, the default masked formulation remains slightly better overall at 97.6\%. These results suggest that explicitly conditioning actions on predicted future features is effective.

\begin{table}[htbp]
\centering
\captionof{table}{\small \textbf{Effect of Action-World Information Flow and Future-Target View.} LIBERO success rates (\%) for \name, \name (Causal), and \name (Joint) action-world coupling using wrist, agent, or dual-camera future-feature supervision. The current visual context is fixed to both cameras for all variants. Best results are \textbf{bold}, and second-best results are \underline{underlined}.}
\label{tab:coupling_camera_view}
\resizebox{0.45\textwidth}{!}{%
\begin{tabular}{lc|cccc|c}
\toprule
\rowcolor{gray!20} \textbf{Coupling} & \textbf{View} & \textbf{Spatial} $\uparrow$ & \textbf{Object} $\uparrow$ & \textbf{Goal} $\uparrow$ & \textbf{Long} $\uparrow$ & \textbf{Average} $\uparrow$ \\
\midrule
\small \textbf{Action Only} & --  & 93.2\% & 97.6\% & 95.4\% & 90.4\% & 94.2\% \\
\midrule
\multirow{3}{*}{\small \textbf{\name}} &\small \textbf{Wrist} & \textbf{98.0}\% & 98.6\% & 96.4\% & 94.6\% & 96.9\% \\
& \small \textbf{Agent} & 95.2\% & 99.2\% & 95.8\% & 95.0\% & 96.3\% \\ 
& \small \textbf{Both} & \textbf{98.0}\% & \textbf{99.6}\% & \underline{97.4}\% & 95.4\% & \textbf{97.6}\% \\
\midrule
\multirow{3}{*}{\small \shortstack{\textbf{\name}\\\textbf{(Causal)}}} & \small \textbf{Wrist} & 97.6\% & \underline{99.4}\% & 96.4\% & \textbf{96.4}\% & \underline{97.4}\% \\
& \small \textbf{Agent} & \underline{97.8}\% & 98.6\% & \textbf{97.8}\% & 95.4\% & \underline{97.4}\% \\
& \small \textbf{Both} & 96.8\% & 98.8\% & 97.0\% & \underline{95.8}\% & 97.1\% \\
\midrule
\multirow{3}{*}{\small \shortstack{\textbf{\name}\\\textbf{(Joint)}}} &\small \textbf{Wrist} & 95.2\% & 98.8\% & \underline{97.4}\% & \underline{95.8}\% & 96.8\% \\
&\small \textbf{Agent} & 94.4\% & 97.4\% & 97.2\% & 94.8\% & 96.0\% \\
&\small \textbf{Both} & 95.8\% & 98.0\% & 95.8\% & 93.2\% & 95.7\% \\
\bottomrule
\end{tabular}}
\end{table}

\begin{figure*}[htbp]
    \centering
    \includegraphics[width=0.95\linewidth]{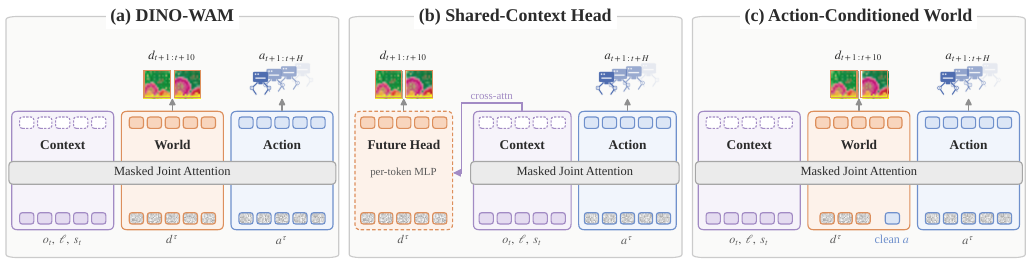}
    \captionof{figure}{\small\textbf{Architectures for Future-Feature Supervision.} (a) \name uses a dedicated world expert to predict future DINO features in parallel with the action expert. Both streams attend to the shared context and within their own modality, but cannot attend to one another. (b) Shared-Context Future Head removes the dedicated world expert and instead predicts the same future DINO targets using a lightweight context-conditioned prediction head. (c) Action-Conditioned World retains the dedicated world expert but additionally conditions future prediction on clean ground-truth action tokens during training, while the action expert remains independent of the world stream. In all three variants, future-feature prediction is used only during training and the deployed policy generates actions from the current context without future-feature tokens.}
    \label{fig:supervision_model}
\end{figure*}

\begin{table*}[htbp]
\centering
\captionof{table}{\small \textbf{Multi-seed evaluation across LIBERO, LIBERO-Plus, and LIBERO-PRO.} Success rates (\%) for Action Only, \name and \name (Joint) are reported as mean $\pm$ standard deviation over three independent training seeds. Best results are \textbf{bold}, and second-best results are \underline{underlined}.}
\label{tab:multiseed_all}
\setlength{\tabcolsep}{3.5pt}
\renewcommand{\arraystretch}{1.1}
\resizebox{\textwidth}{!}{
\begin{tabular}{l|ccccc|cccccccc|ccccc}
\toprule
\rowcolor{gray!20} & \multicolumn{5}{c|}{\textbf{LIBERO}} & \multicolumn{8}{c|}{\textbf{LIBERO-Plus}} & \multicolumn{5}{c}{\textbf{LIBERO-PRO}} \\
\rowcolor{gray!20}
\textbf{Method} & \textbf{Spatial} $\uparrow$ & \textbf{Object} $\uparrow$& \textbf{Goal} $\uparrow$& \textbf{Long} $\uparrow$& \textbf{Avg.} $\uparrow$& \textbf{Camera} $\uparrow$& \textbf{Robot} $\uparrow$& \textbf{Lang.} $\uparrow$& \textbf{Light} $\uparrow$& \textbf{Backg.} $\uparrow$& \textbf{Noise} $\uparrow$& \textbf{Layout} $\uparrow$& \textbf{Avg.} $\uparrow$& \textbf{Spatial} $\uparrow$& \textbf{Object} $\uparrow$& \textbf{Goal} $\uparrow$& \textbf{Long} $\uparrow$& \textbf{Avg.} $\uparrow$ \\
\midrule
\textbf{Action Only} &\small 94.3$\pm$1.1\% &\small 97.9$\pm$0.4\% &\small 95.9$\pm$1.5\% &\small 92.3$\pm$1.7\% &\small 95.1$\pm$0.8\% &\small \underline{80.0$\pm$1.7}\% &\small 36.8$\pm$3.0\% &\small 71.9$\pm$1.4\% &\small 90.0$\pm$1.6\% &\small \underline{57.2$\pm$4.3}\% &\small \underline{72.2$\pm$7.6}\% &\small 72.7$\pm$1.4\% &\small \underline{68.4$\pm$1.6}\% &\small 55.5$\pm$0.6\% &\small \underline{38.3$\pm$1.1}\% &\small 53.7$\pm$0.4\% &\small 36.5$\pm$1.7\% &\small 46.0$\pm$0.6\% \\

\textbf{\name} &\small \textbf{97.3$\pm$1.4}\% &\small \textbf{99.3$\pm$0.4}\% &\small \textbf{96.7$\pm$0.7}\% &\small \textbf{95.1$\pm$0.2}\% &\small \textbf{97.1$\pm$0.5}\% &\small \textbf{89.6$\pm$2.2}\% &\small \underline{37.7$\pm$1.3}\% &\small \underline{77.9$\pm$1.2}\% &\small \textbf{92.7$\pm$0.9}\% &\small \textbf{66.5$\pm$1.2}\% &\small \textbf{81.4$\pm$5.3}\% &\small \textbf{77.7$\pm$0.9}\% &\small \textbf{74.5$\pm$1.2}\% &\small \textbf{56.8$\pm$0.6}\% &\small \textbf{38.9$\pm$0.4}\% &\small \textbf{54.7$\pm$1.2}\% &\small \textbf{39.7$\pm$0.6}\% &\small \textbf{47.5$\pm$0.5}\%\\

\textbf{\name{} (Joint)}&\small \underline{95.0$\pm$1.1}\% &\small \underline{98.7$\pm$0.6}\% &\small \underline{96.2$\pm$0.7}\% &\small \underline{94.2$\pm$1.1}\% &\small \underline{96.0$\pm$0.3}\% &\small 75.1$\pm$0.6\% &\small \textbf{40.7$\pm$2.9}\% &\small \textbf{79.7$\pm$2.9}\% &\small \underline{91.6$\pm$1.5}\% &\small 39.9$\pm$4.4\% &\small 63.5$\pm$4.1\% &\small \underline{75.0$\pm$0.3}\% &\small 66.7$\pm$1.7\% &\small \underline{55.7$\pm$0.7}\% &\small 37.8$\pm$0.4\% &\small \underline{54.3$\pm$0.4}\% &\small \underline{36.8$\pm$0.7\%} &\small \underline{46.2$\pm$0.5}\% \\
\bottomrule
\end{tabular}}
\end{table*}

\subsection{What Aspect of Future Supervision Matters?}
We further isolate which component of future-feature supervision is responsible for the policy improvement. In addition to the action-only baseline and the default masked \name formulation, we evaluate two architectural controls, illustrated in Figure~\ref{fig:supervision_model}. The \emph{Shared-Context Future Head} removes the dedicated world transformer expert and instead predicts the same future DINO features from the shared context representation through a lightweight context-conditioned prediction head. This control preserves the same future targets and action-only inference path, testing whether a dedicated world stream is necessary beyond providing auxiliary future supervision. The \emph{Action-Conditioned World} variant retains the dedicated world expert but additionally conditions future prediction on the clean ground-truth action chunk during training. The action stream remains independent of the world stream, and the world branch is removed at inference. This variant therefore tests whether explicitly learning action-conditioned future dynamics provides additional benefit over context-conditioned future prediction.

\begin{table}[htbp]
\centering
\captionof{table}{\small \textbf{What Aspect of Future Supervision Matters?} LIBERO success rates (\%) for the action-only baseline, different action-world coupling strategies, and alternative future-supervision architectures. Best results are \textbf{bold}, and second-best results are \underline{underlined}.}
\label{tab:future_supervision_arch}
\resizebox{0.95\linewidth}{!}{
\begin{tabular}{l|cccc|c}
\toprule
\rowcolor{gray!20} \small \textbf{Method} & \small \textbf{Spatial} $\uparrow$ & \small \textbf{Object} $\uparrow$ & \small \textbf{Goal} $\uparrow$ & \small \textbf{Long} $\uparrow$ & \small \textbf{Average} $\uparrow$ \\
\midrule
\small \textbf{Action Only} & 93.2\% & 97.6\% & 95.4\% & 90.4\% & 94.2\% \\
\midrule
\small \textbf{\name{} (Joint)} & 95.8\% & 98.0\% & 95.8\% & 93.2\% & 95.7\% \\
\small \textbf{\name{} (Causal)} & 96.8\% & 98.8\% & 97.0\% & 95.8\% & 97.1\% \\
\small \textbf{\name} & \underline{98.0}\% & \textbf{99.6}\% & 97.4\% & 95.4\% & \underline{97.6}\% \\
\midrule
\small \textbf{Shared-Context Future Head} & 97.8\% & 99.2\% & \textbf{97.6}\% & \textbf{96.6}\% & \textbf{97.8}\% \\
\small \textbf{Action-Conditioned World} & 97.2\% & 98.2\% & 97.2\% & \underline{95.8}\% & 97.1\% \\
\bottomrule
\end{tabular}}
\end{table}

Table~\ref{tab:future_supervision_arch} compares these controls with the action-only, joint, causal, and masked formulations. For the coupling variants, we fix future-feature supervision to the same dual-camera setting to isolate the effect of architectural information flow. All future-supervised variants outperform the action-only baseline, confirming that future prediction provides a useful training signal. Notably, the Shared-Context Future Head achieves 97.8\% average success, comparable to the default masked model at 97.6\%, despite removing the dedicated world transformer. This indicates that much of the benefit can be obtained by using future-feature prediction to shape the shared representation, rather than requiring a dedicated generative world stream. The Action-Conditioned World variant reaches 97.1\%, showing that conditioning future prediction on clean actions does not provide an additional improvement in this setting. Together with the causal and joint results, these controls suggest that the primary benefit of future prediction arises from auxiliary predictive supervision.

\setlength{\tabcolsep}{3pt}
\begin{table*}[!t]
\centering
\captionof{table}{\small \textbf{Ablation Studies: Zero-shot Results on LIBERO-Plus.} Success rates (\%) are reported under seven perturbation types. ``Lang.'' denotes Language and ``Backg.'' denotes Background. Best results within each ablation block are \textbf{bold}, and second-best results are \underline{underlined}.}
\label{tab:liberoplus_supp}
\scriptsize
\renewcommand{\arraystretch}{1.0}
\resizebox{0.9\textwidth}{!}{%
\begin{tabular}{l|ccccccc|c}
\toprule
\rowcolor{gray!20} \textbf{Method} & \textbf{Camera} $\uparrow$ & \textbf{Robot} $\uparrow$ & \textbf{Lang.} $\uparrow$ & \textbf{Light} $\uparrow$ & \textbf{Backg.} $\uparrow$ & \textbf{Noise} $\uparrow$ & \textbf{Layout} $\uparrow$ & \textbf{Overall} $\uparrow$ \\
\midrule
\rowcolor{dinowamblue}
\multicolumn{9}{c}{\textit{Effect of Action--World Attention}} \\
\textbf{Action Only} & \underline{78.8} & \textbf{40.3} & 71.1 & 89.7 & \underline{52.5} & \underline{63.8} & 71.3 & \underline{66.5} \\
\textbf{\name{} (Joint)} & 75.1 & \underline{38.3} & \textbf{79.8} & \underline{90.7} & 34.8 & 58.8 & \underline{75.3} & 65.0 \\
\textbf{\name{}} & \textbf{87.3} & 36.3 & \underline{79.2} & \textbf{93.1} & \textbf{65.7} & \textbf{83.3} & \textbf{77.6} & \textbf{74.4} \\

\midrule
\rowcolor{dinowamblue} 
\multicolumn{9}{c}{\textit{What Aspect of Future Supervision Matters?}} \\
\textbf{\name{}} & \underline{87.3} & 36.3 & 79.2 & \underline{93.1} & \underline{65.7} & 83.3 & 77.6 & 74.4 \\
\textbf{Shared-Context Future Head} & \textbf{89.1} & \textbf{41.6} & \textbf{79.8} & \textbf{94.0} & \textbf{69.9} & \textbf{84.4} & \textbf{78.9} & \textbf{76.5} \\
\textbf{Action-Conditioned World} & 86.7 & \underline{40.1} & \underline{79.3} & 92.6 & \textbf{69.9} & \underline{83.4} & \underline{78.2} & \underline{75.4} \\

\midrule
\rowcolor{dinowamblue} 
\multicolumn{9}{c}{\textit{Absolute vs. Residual Feature Supervision}} \\
\textbf{\name{} (Absolute)} & \underline{87.3} & \underline{36.3} & \underline{79.2} & \underline{93.1} & \textbf{65.7} & \textbf{83.3} & \textbf{77.6} & \textbf{74.4} \\
\textbf{\name{} (Residual)} & \textbf{89.6} & 34.6 & 78.9 & \textbf{94.1} & \underline{64.2} & \underline{77.1} & \underline{76.9} & \underline{73.3} \\
\textbf{\name{} (Joint, Absolute)} & 75.1 & \textbf{38.3} & \textbf{79.8} & 90.7 & 34.8 & 58.8 & 75.3 & 65.0 \\
\textbf{\name{} (Joint, Residual)} & 62.4 & 35.4 & 70.6 & 79.1 & 30.5 & 42.2 & 64.9 & 55.1 \\

\midrule
\rowcolor{dinowamblue} 
\multicolumn{9}{c}{\textit{Dense vs. Sparse Feature Supervision}} \\
\textbf{\name{} (Dense)} & \underline{87.3} & 36.3 & \underline{79.2} & \underline{93.1} & \textbf{65.7} & \textbf{83.3} & \textbf{77.6} & \textbf{74.4} \\
\textbf{\name{} (Sparse)} & \textbf{90.5} & 37.2 & 77.6 & \textbf{94.8} & \underline{65.1} & \underline{81.4} & \underline{75.9} & \textbf{74.4} \\
\textbf{\name{} (Joint, Dense)} & 75.1 & \underline{38.3} & \textbf{79.8} & 90.7 & 34.8 & 58.8 & 75.3 & \underline{65.0} \\
\textbf{\name{} (Joint, Sparse)} & 43.2 & \textbf{39.0} & 77.9 & 85.6 & 35.6 & 45.4 & 65.3 & 55.6 \\

\midrule
\rowcolor{dinowamblue} 
\multicolumn{9}{c}{\textit{Effect of Future-Target Camera View and Coupling}} \\
\textbf{\name{} (Wrist)} & 86.5 & 34.5 & 76.5 & \textbf{94.5} & 60.9 & 79.1 & 76.1 & 72.3 \\
\textbf{\name{} (Agent)} & 83.8 & 38.2 & 75.9 & \underline{94.4} & 64.0 & 76.0 & 76.3 & 72.2 \\
\textbf{\name{} (Both)} & \underline{87.3} & 36.3 & 79.2 & 93.1 & \textbf{65.7} & \textbf{83.3} & \underline{77.6} & \textbf{74.4} \\
\textbf{\name{} (Joint, Wrist)} & 83.4 & 40.1 & \textbf{81.0} & 90.7 & 48.0 & \underline{81.7} & \textbf{77.8} & 72.3 \\
\textbf{\name{} (Joint, Agent)} & 64.4 & 38.4 & \underline{80.2} & 89.6 & 49.2 & 60.8 & 72.3 & 64.7 \\
\textbf{\name{} (Joint, Both)} & 75.1 & 38.3 & 79.8 & 90.7 & 34.8 & 58.8 & 75.3 & 65.0 \\
\textbf{\name{} (Causal, Wrist)} & \textbf{87.4} & 37.9 & 80.1 & 92.2 & 55.4 & 78.0 & 77.5 & \underline{72.7} \\
\textbf{\name{} (Causal, Agent)} & 72.0 & \textbf{42.2} & 77.4 & 94.0 & \underline{65.1} & 60.0 & 72.5 & 68.2 \\
\textbf{\name{} (Causal, Both)} & 84.3 & \underline{41.9} & 76.4 & 91.9 & 48.0 & 76.0 & 75.7 & 70.9 \\

\midrule
\rowcolor{dinowamblue} 
\multicolumn{9}{c}{\textit{Effect of Visual Encoder Choice}} \\
\textbf{\name{} (DINOv2)} & 87.3 & 36.3 & 79.2 & \underline{93.1} & 65.7 & \underline{83.3} & \underline{77.6} & 74.4 \\
\textbf{\name{} (SigLIP)} & \textbf{90.1} & \textbf{49.6} & 75.8 & \textbf{95.0} & \underline{79.8} & 79.3 & 75.5 & \textbf{77.2} \\
\textbf{\name{} (V-JEPA 2)} & \underline{88.0} & 29.4 & 66.2 & 91.5 & \textbf{82.2} & \textbf{85.4} & 73.0 & 72.7 \\
\textbf{\name{} (Joint, DINOv2)} & 75.1 & 38.3 & \underline{79.8} & 90.7 & 34.8 & 58.8 & 75.3 & 65.0 \\
\textbf{\name{} (Joint, SigLIP)} & 86.0 & \underline{48.6} & \textbf{82.6} & 90.7 & 74.8 & 79.6 & \textbf{78.8} & \underline{76.9} \\
\textbf{\name{} (Joint, V-JEPA 2)} & 58.6 & 29.9 & 68.4 & 88.7 & 71.4 & 63.5 & 69.6 & 62.9 \\

\midrule
\rowcolor{dinowamblue} 
\multicolumn{9}{c}{\textit{Effect of DINO Context Layers}} \\
\textbf{\name{}} & \underline{87.3} & \textbf{36.3} & \textbf{79.2} & \textbf{93.1} & \textbf{65.7} & \textbf{83.3} & \textbf{77.6} & \textbf{74.4} \\
\textbf{\name{} (Final Layer)} & \textbf{88.3} & \underline{34.6} & \underline{77.6} & \textbf{93.1} & \underline{62.7} & \underline{79.3} & \underline{74.6} & \underline{72.6} \\

\midrule
\rowcolor{dinowamblue} 
\multicolumn{9}{c}{\textit{Future-Feature Token Budget}} \\
\textbf{\name{}} & \underline{87.3} & \underline{36.3} & \textbf{79.2} & \textbf{93.1} & \underline{65.7} & \textbf{83.3} & \textbf{77.6} & \underline{74.4} \\
\textbf{\name{} (8 Tokens/View)} & \textbf{90.6} & \textbf{38.4} & \underline{78.1} & \underline{92.8} & \textbf{66.6} & \underline{80.8} & \underline{77.1} & \textbf{74.7} \\

\midrule
\rowcolor{dinowamblue} 
\multicolumn{9}{c}{\textit{Effect of Future-Feature Loss Weight}} \\
\textbf{\name{} ($\lambda_{\mathrm{world}}=0.25$)} & 85.4 & 35.2 & 75.6 & 92.9 & 58.6 & 65.5 & 75.1 & 69.4 \\
\textbf{\name{} ($\lambda_{\mathrm{world}}=1.0$)} & \underline{87.3} & \underline{36.3} & \textbf{79.2} & \underline{93.1} & \underline{65.7} & \textbf{83.3} & \underline{77.6} & \underline{74.4} \\
\textbf{\name{} ($\lambda_{\mathrm{world}}=2.0$)} & \textbf{90.2} & \textbf{38.1} & \underline{78.5} & \textbf{93.9} & \textbf{66.5} & \underline{77.8} & \textbf{78.8} & \textbf{74.5} \\

\midrule
\rowcolor{dinowamblue} 
\multicolumn{9}{c}{\textit{Effect of DINO Context Feature Fusion}} \\
\textbf{\name{} (Learned Fusion)} & \underline{87.3} & \textbf{36.3} & \textbf{79.2} & \textbf{93.1} & \textbf{65.7} & \underline{83.3} & \textbf{77.6} & \textbf{74.4} \\
\textbf{\name{} (Mean Fusion)} & 86.8 & 30.0 & 74.0 & \underline{93.0} & 56.3 & \textbf{83.9} & \underline{75.4} & 71.3 \\
\textbf{\name{} (Final Layer)} & \textbf{88.3} & \underline{34.6} & \underline{77.6} & \textbf{93.1} & \underline{62.7} & 79.3 & 74.6 & \underline{72.6} \\
\bottomrule
\end{tabular}%
}
\end{table*}

\begin{table*}[t]
\centering
\captionof{table}{\small \textbf{Zero-Shot Robustness on LIBERO-PRO.} Success rates (\%) are reported across the four LIBERO suites, with each suite averaged over five LIBERO-PRO perturbations. Best results are \textbf{bold}, and second-best results are \underline{underlined}.}
\label{tab:liberopro}
\renewcommand{\arraystretch}{1.0}
\begin{tabular}{l|cc|cccc|c}
\toprule
\rowcolor{gray!20} \small \textbf{Method} & \small \textbf{Params} & \small \textbf{Embodied PT.} &\small \textbf{Spatial} $\uparrow$ &\small \textbf{Object} $\uparrow$ &\small \textbf{Goal} $\uparrow$ &\small \textbf{Long} $\uparrow$ &\small \textbf{Average} $\uparrow$ \\
\midrule
\rowcolor{dinowamblue} \multicolumn{8}{@{}l}{\emph{Published Methods}} \\
\small OpenVLA~\cite{kim2024openvla} &\small 7B&\small \cmark&\small  40.5\% &\small  32.0\% &\small  25.0\% &\small  20.5\% &\small  29.5\% \\
\small $\pi_0$~\cite{black2024pi_0} &\small 3B&\small \cmark& \small 52.0\% &\small 33.2\% &\small 54.6\% &\small 33.2\% &\small 43.2\% \\
\small $\pi_{0.5}$~\cite{intelligence2025pi_} &\small 3B&\small \cmark&\small 55.6\% &\small \underline{52.2}\% &\small 48.0\% &\small \textbf{57.0}\% &\small \underline{53.0}\% \\
\small Fast-WAM~\cite{yuan2026fast} &\small 6B&\small \xmark& \small \underline{57.6}\% &\small 49.2\% &\small \underline{58.1}\% &\small 40.8\% &\small 51.4\% \\
\small Faster-WAM~\cite{zhao2026faster} &\small 5B&\small \xmark&\small \textbf{59.6}\% &\small \textbf{52.8}\% &\small \textbf{59.6}\% &\small \underline{45.8}\% &\small \textbf{54.5}\% \\
\midrule
\rowcolor{dinowamblue} \multicolumn{8}{@{}l}{\emph{\name Variants}} \\
\small Action Only &\small 0.4B&\small \xmark & \small 54.9\% &\small 37.8\% &\small 54.1\% &\small 34.5\% &\small 45.3\% \\
\small \name (Joint) &\small 0.4B&\small \xmark & \small 55.6\% &\small 37.4\% &\small 53.8\% &\small 36.0\% &\small 45.7\% \\
\small \name (Causal) &\small 0.4B&\small \xmark & \small 56.8\% &\small 37.6\% &\small 56.2\% &\small 39.0\% &\small 47.4\% \\
\small \name &\small 0.4B&\small \xmark & \small 57.0\% &\small 38.8\% &\small 55.2\% &\small 40.2\% &\small 47.8\% \\
\bottomrule
\end{tabular}
\end{table*}

\begin{table}[htbp]
\centering
\captionof{table}{\small \textbf{Inference efficiency of future-supervised variants.}
Batch-one latency for 20-step action sampling on an RTX 4080 SUPER. ``Tokens/step'' denotes the number of tokens processed at each denoising step, and ``Active Params'' denotes parameters executed during inference. Timing excludes frozen visual and language encoders and simulator execution. ``SCFH'' and ``ACW'' denote Shared-Context Future Head and Action-Conditioned World, respectively.}
\label{tab:inference_cost}
\setlength{\tabcolsep}{2pt}
\resizebox{0.98\linewidth}{!}{
\begin{tabular}{l|ccc}
\toprule
\rowcolor{gray!20} \small \textbf{Method} & \small \textbf{Tokens/step} & \small \textbf{Active Params (M)} &\small \textbf{Latency (ms)} \\
\midrule
\small \name& 44  & 228.8 & 91.8 \\
\small \name{} (Joint)& 364 & 343.4 & 187.6 \\
\small \name{} (Causal)& 364 & 343.4 & 194.7 \\
\small SCFH & 44  & 228.8 & 91.9 \\
\small ACW & 44  & 228.8 & 93.3 \\
\bottomrule
\end{tabular}}
\end{table}

\section{Multi-Seed Evaluation}
To assess the stability of the main findings across training randomness, we repeat the Action Only, \name, and \name{} (Joint) experiments using three independent training seeds and report mean $\pm$ standard deviation in Table~\ref{tab:multiseed_all}. The default \name formulation consistently achieves the highest mean performance across all three benchmarks, reaching 97.1$\pm$0.5\% on LIBERO, 74.5$\pm$1.2\% on LIBERO-Plus, and 47.5$\pm$0.5\% on LIBERO-PRO. In comparison, the action-only baseline obtains 95.1$\pm$0.8\%, 68.4$\pm$1.6\%, and 46.0$\pm$0.6\%, respectively.

The jointly coupled variant achieves 96.0$\pm$0.3\% on LIBERO, 66.7$\pm$1.7\% on LIBERO-Plus, and 46.2$\pm$0.5\% on LIBERO-PRO. While it improves over the action-only baseline on standard LIBERO, it does not provide a consistent advantage under the two distribution-shift benchmarks. In contrast, the masked \name formulation maintains higher mean performance than both alternatives across all three evaluations.

\section{Robustness Under Distribution Shift}
\subsection{LIBERO-Plus}
We further evaluate whether the design trends observed on standard LIBERO remain consistent under zero-shot distribution shifts. Table~\ref{tab:liberoplus_supp} reports LIBERO-Plus performance for the main ablation variants across camera, robot, language, lighting, background, noise, and layout perturbations.

Overall, the masked \name formulation remains the strongest action-world coupling design, outperforming both the action-only and joint variants. Among alternative future-supervision architectures, the Shared-Context Future Head achieves the highest overall robustness, suggesting that auxiliary future-feature supervision can substantially improve the shared representation even without a dedicated world transformer. Sparse supervision remains competitive with dense supervision, while residual targets substantially degrade the jointly coupled model. Future-target viewpoint interacts with the coupling structure, with dual-camera supervision performing best for the masked formulation and different views being preferred by the joint and causal variants.

Finally, performance is relatively insensitive to moderate changes in future-token budget and future-feature loss weighting, while learned multi-layer DINO context fusion remains more robust than simpler context representations.

\subsection{LIBERO-PRO}
We further evaluate whether the trends observed on LIBERO and LIBERO-Plus extend to LIBERO-PRO, which introduces additional zero-shot perturbations across the four LIBERO suites. Table~\ref{tab:liberopro} compares the main \name variants with representative published VLA and WAM baselines.

Although the published large-scale models achieve higher absolute performance, the comparison among our controlled variants reveals a consistent trend. The action-only baseline reaches 45.3\% average success, while direct joint action--world generation provides only a small improvement to 45.7\%. Introducing directional world-to-action coupling increases performance to 47.4\%, and the default masked \name formulation reaches 47.8\%.

These results reinforce the conclusions from LIBERO and LIBERO-Plus. Future-feature prediction provides a consistent improvement over action-only training, but stronger coupling between the world and action streams is not required for robust policy learning. The similar performance of the masked world expert and the lightweight Shared-Context Future Head further suggests that much of the benefit arises from predictive supervision shaping the shared representation rather than from explicitly generating or consuming future features at deployment.

\section{Inference Efficiency}
Table~\ref{tab:inference_cost} compares the deployment cost of the future-supervised variants. The default \name formulation removes the world stream at inference, processing only 44 tokens per denoising step with a latency of 91.8\,ms. In contrast, the Joint and Causal variants retain future-feature tokens during action generation, increasing the token count to 364 and the latency to 187.6\,ms and 194.7\,ms, respectively. SCFH and ACW also discard their auxiliary prediction pathways at deployment and therefore retain essentially the same inference cost as the default \name model. These results show that future-feature supervision can improve policy learning without requiring explicit future generation at test time.

\end{document}